\documentclass[sigconf]{acmart}

\usepackage{makecell}
\usepackage{fancybox}
\usepackage{framed}
\usepackage{tcolorbox}
\usepackage{algorithm}
\usepackage{algpseudocode}
\usepackage{tikz}
\usepackage{graphicx} 
\usepackage{booktabs}
\usepackage{multirow}
\usepackage{subcaption}  % For subfigures
\usepackage[normalem]{ulem}
\usepackage{pifont}
\useunder{\uline}{\ul}{}
\AtBeginDocument{%
  }

\copyrightyear{2025}
\acmYear{2025}
\setcopyright{acmlicensed}\acmConference[KDD '25]{Proceedings of the 31st ACM SIGKDD Conference on Knowledge Discovery and Data Mining V.2}{August 3--7, 2025}{Toronto, ON, Canada}
\acmBooktitle{Proceedings of the 31st ACM SIGKDD Conference on Knowledge Discovery and Data Mining V.2 (KDD '25), August 3--7, 2025, Toronto, ON, Canada}
\acmDOI{10.1145/3711896.3736872}
\acmISBN{979-8-4007-1454-2/2025/08}
\begin{document}

%%
%% The "title" command has an optional parameter,
%% allowing the author to define a "short title" to be used in page headers.\

% \title{Multi-Model Financial Forecasting Considering Time Series Pattern and Salient Policy Textual Information}

\title[CAMEF: Causal-Augmented Multi-Modality Financial Forecasting]{CAMEF: Causal-Augmented Multi-Modality Event-Driven Financial Forecasting by Integrating Time Series Patterns and Salient Macroeconomic Announcements}

%% Event-Driven Financial Forecasting with Causal Data Augmentation: A Multi-Modality Approach Using Time Series and Macroeconomic Announcements

%%
%% The "author" command and its associated commands are used to define
%% the authors and their affiliations.
%% Of note is the shared affiliation of the first two authors, and the
%% "authornote" and "authornotemark" commands
%% used to denote shared contribution to the research.

% \author{Anonymous Authors}

% % \authornote{Both authors contributed equally to this research.}
% % \email{trovato@corporation.com}
% % \orcid{1234-5678-9012}

\author{Yang Zhang}
\authornote{Corresponding authors.}
\affiliation{%
  \institution{Southwestern University of Finance and Economics}
  \city{Chengdu}
  \country{China}
}
\email{zhang.yang.r54@kyoto-u.jp}

\author{Wenbo Yang}
\affiliation{%
  \institution{Southwestern University of Finance and Economics}
  \city{Chengdu}
  \country{China}
}
\email{223081200030@smail.swufe.edu.cn}

\author{Jun Wang}
\authornotemark[1]  % Use the same footnote mark as Yang Zhang
\affiliation{%
  \institution{Southwestern University of Finance and Economics}
  \city{Chengdu}
  \country{China}
}
\email{wangjun1987@swufe.edu.cn}

\author{Qiang Ma}
\affiliation{%
  \institution{Kyoto Institute of Technology}
  \city{Kyoto}
  \country{Japan}
}
\email{qinag@kit.ac.jp}

\author{Jie Xiong}
\affiliation{%
  \institution{Southwestern University of Finance and Economics}
  \city{Chengdu}
  \country{China}
}
\email{xiongjie@swufe.edu.cn}

% \renewcommand{\shortauthors}{CAMEF}

%%
%% The abstract is a short summary of the work to be presented in the
%% article.
\begin{abstract}

Accurately forecasting the impact of macroeconomic events is critical for investors and policymakers. Salient events like monetary policy decisions and employment reports often trigger market movements by shaping expectations of economic growth and risk, thereby establishing causal relationships between events and market behavior. Existing forecasting methods typically focus either on textual analysis or time-series modeling, but fail to capture the multi-modal nature of financial markets and the causal relationship between events and price movements. To address these gaps, we propose \textbf{CAMEF} (Causal-Augmented Multi-Modality Event-Driven Financial Forecasting), a multi-modality framework that effectively integrates textual and time-series data with a causal learning mechanism and an LLM-based counterfactual event augmentation technique for causal-enhanced financial forecasting. Our contributions include: (1) a multi-modal framework that captures causal relationships between policy texts and historical price data; (2) a new financial dataset with six types of macroeconomic releases from 2008 to April 2024, and high-frequency real trading data for five key U.S. financial assets; and (3) an LLM-based counterfactual event augmentation strategy. We compare CAMEF to state-of-the-art transformer-based time-series and multi-modal baselines, and perform ablation studies to validate the effectiveness of the causal learning mechanism and event types.

\end{abstract}

%%
%% The code below is generated by the tool at http://dl.acm.org/ccs.cfm.
%% Please copy and paste the code instead of the example below.
%%
\begin{CCSXML}
<ccs2012>
   <concept>
       <concept_id>10010405.10010455.10010460</concept_id>
       <concept_desc>Applied computing~Economics</concept_desc>
       <concept_significance>300</concept_significance>
       </concept>
   <concept>
       <concept_id>10010147.10010257.10010293.10010294</concept_id>
       <concept_desc>Computing methodologies~Neural networks</concept_desc>
       <concept_significance>500</concept_significance>
       </concept>
 </ccs2012>
\end{CCSXML}

\ccsdesc[300]{Applied computing~Economics}
\ccsdesc[500]{Computing methodologies~Neural networks}

% \ccsdesc[500]{Do Not Use This Code~Generate the Correct Terms for Your Paper}
% \ccsdesc[300]{Do Not Use This Code~Generate the Correct Terms for Your Paper}
% \ccsdesc{Do Not Use This Code~Generate the Correct Terms for Your Paper}
% \ccsdesc[100]{Do Not Use This Code~Generate the Correct Terms for Your Paper}

%%
%% Keywords. The author(s) should pick words that accurately describe
%% the work being presented. Separate the keywords with commas.
\keywords{Multimodal learning, Causal Learning, Financial dataset, Time-series Forecasting}
%% A "teaser" image appears between the author and affiliation
%% information and the body of the document, and typically spans the
%% page.

% \received{20 February 2007}
% \received[revised]{12 March 2009}
% \received[accepted]{5 June 2009}

%%
%% This command processes the author and affiliation and title
%% information and builds the first part of the formatted document.
\maketitle

\section{Introduction}

% Fama's empirical work on the Efficient Market Hypothesis (EMH) provides evidence that financial markets are informationally efficient \cite{1731c543-0a2e-3dd1-85c2-3ffc09a485a7}, meaning asset prices incorporate all available market information. 

\begin{figure*}[t]
    \centering
    \includegraphics[width=0.8\linewidth]{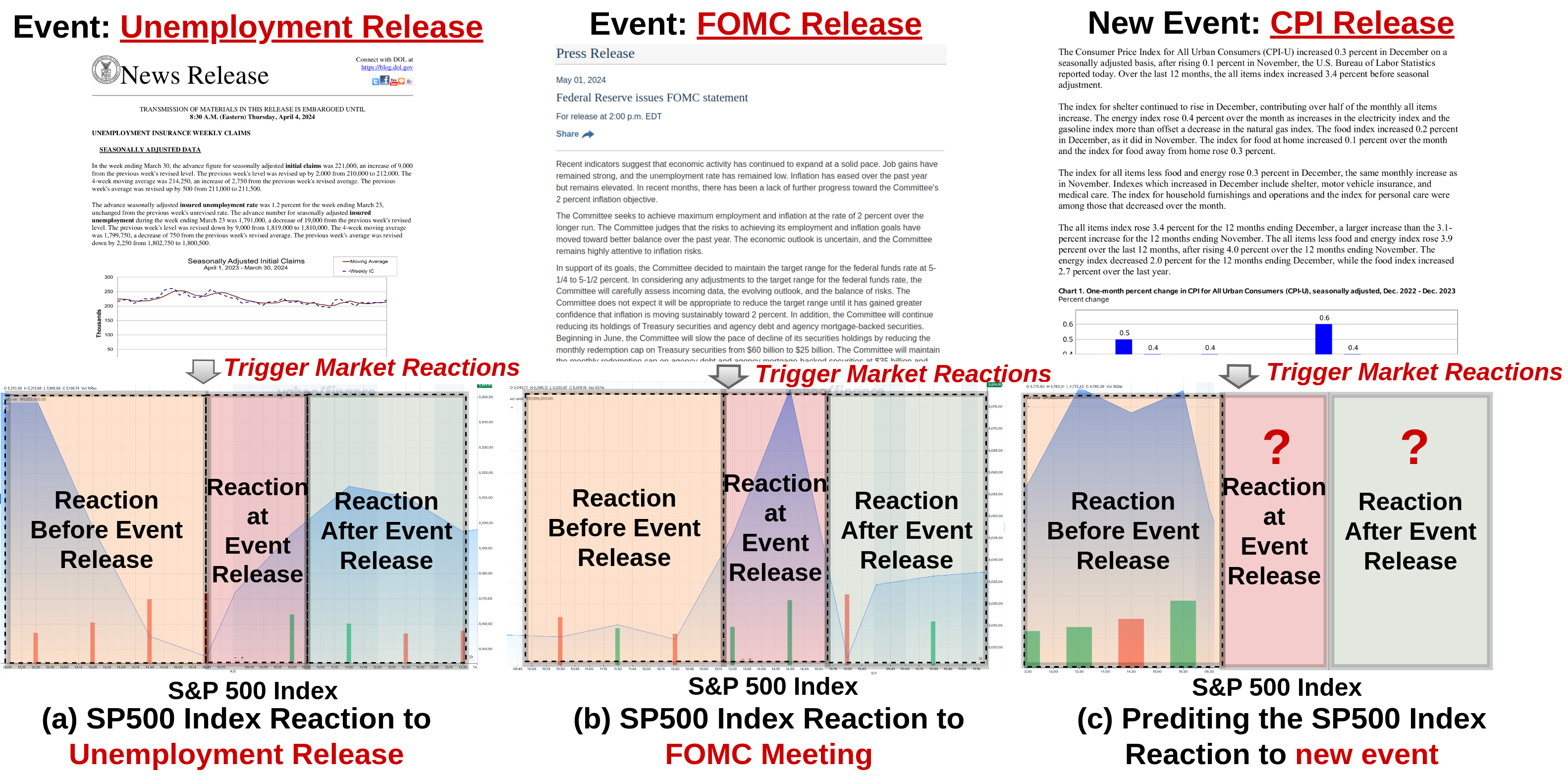}
    \caption{Event-Driven Forecasting Examples: (a) Market reaction to employment insurance release; (b) Market reaction during FOMC meeting; (c) Forecasting market reactions to future events.}
    \label{fig:multi-model-demo}
\end{figure*}

The prices of financial assets reflect all available information, according to Fama's Efficient Market Theory \cite{1731c543-0a2e-3dd1-85c2-3ffc09a485a7,78bd1d1d-3f88-3c53-ad25-c8ac73767958}. Major financial releases from government sectors often trigger market movements by shaping investors' expectations and evaluations of economic conditions, asset growth potential, and associated risks. For example, during the FOMC meeting on March 16, 2020, the Fed’s emergency rate cut to 0-0.25\% sharply altered investors' economic outlook, resulting in a massive sell-off. Major indices, including the S\&P 500, NASDAQ, and Dow Jones, dropped by over 10\%, marking the steepest single-day decline since 1987 \cite{cnbc2021}. These salient macroeconomic events cause reactions in financial assets, establishing causal relationships between events and financial assets. Figure \ref{fig:multi-model-demo} illustrates multiple types of events that cause financial market reactions. Therefore, \textbf{accurately forecasting the causal consequences of the salient macroeconomic releases on financial market is essential, not only to help investors manage risks and maximize returns, but also to provide policymakers with valuable insights for evaluating and refining future policies.}

Previous studies on event-driven forecasting have primarily adopted three lines of methodologies. The first line of approaches utilizes text feature-based models, where language models, ranging from self-crafted RNN-based architectures \cite{10.1145/3155133.3155202,liu2018leveragingfinancialnewsstock,doi:10.1080/14697688.2019.1622314,xu-cohen-2018-stock} to pre-trained transformers \cite{zhou-etal-2021-trade,shah-etal-2023-trillion}, embed sentiment information into text vectors, and then stock movements are predicted as a binary classification task (e.g., hawkish vs. dovish). The second line of methodology focuses on historical time-series data, treating stock price movements as a regression problem \cite{RePEc:wsi:wsbook:6578,https://doi.org/10.1002/jae.3950070512}. Recently, transformer-based architectures have been applied for time-series prediction, including Informer \cite{Zhou_Zhang_Peng_Zhang_Li_Xiong_Zhang_2021}, FedFormer \cite{pmlr-v162-zhou22g}, and AutoFormer \cite{Chen_2021_ICCV}, etc. However, both of these directions typically focus on a single modality, neglecting multi-dimensional information. The third line of research adopts a multi-modality approach, leveraging multiple types of data sources to enhance forecasting performance. For instance, studies like \cite{10.1145/3503161.3548380,ouyang-etal-2024-modal} incorporate textual, video, and audio data from FOMC meetings alongside corresponding market movements. While these approaches show promise for event-driven financial forecasts, they face three major limitations:

\begin{itemize}
    \item \textbf{Data Limitation:} Existing approaches predominantly focus on a single type of event, such as FOMC meetings \cite{shah-etal-2023-trillion,10.1145/3503161.3548380,ouyang-etal-2024-modal}, while neglecting other crucial macroeconomic events like unemployment insurance releases, CPI, PPI, and GDP advance reports. Additionally, many studies rely on daily-based time-series data for financial assets \cite{shah-etal-2023-trillion,10.1145/3503161.3548380,ouyang-etal-2024-modal,RePEc:wsi:wsbook:6578,https://doi.org/10.1002/jae.3950070512,Zhou_Zhang_Peng_Zhang_Li_Xiong_Zhang_2021,pmlr-v162-zhou22g,Chen_2021_ICCV}, which limits their applicability and precision in real-time trading scenarios where high-frequency data is mostly adopted.

    \item \textbf{Modality Limitation:} Most prior studies rely on single-modality analysis, using either textual models \cite{10.1145/3155133.3155202,liu2018leveragingfinancialnewsstock,doi:10.1080/14697688.2019.1622314,xu-cohen-2018-stock,zhou-etal-2021-trade,shah-etal-2023-trillion} or time-series models \cite{RePEc:wsi:wsbook:6578,https://doi.org/10.1002/jae.3950070512,Zhou_Zhang_Peng_Zhang_Li_Xiong_Zhang_2021,pmlr-v162-zhou22g,Chen_2021_ICCV}, which fail to integrate the complementary strengths of both modalities. While some multi-modality approaches have been proposed \cite{10.1145/3503161.3548380,ouyang-etal-2024-modal}, they often lack advanced mechanisms for feature fusion, effective decoding strategies, and causal learning, which are critical for understanding the complex interplay between event texts and market dynamics.

    \item \textbf{Causality Limitation:} Existing methods \cite{10.1145/3503161.3548380,ouyang-etal-2024-modal} fail to incorporate causal reasoning frameworks, overlooking the causal relationships between events and market reactions. Without explicitly modeling these relationships, such approaches cannot fully capture the drivers of financial market behavior, limiting their predictive robustness.
\end{itemize}

% \begin{figure*}[t]
%     \centering
%     \includegraphics[width=0.7\linewidth]{figures/Counterfactual_Generation.drawio.png}
%     \caption{Overview of the Counterfactual Event Generation and Learning Process via Counterfactual Event \textcolor{red}{[changed the figure to be purely examples]}}
%     \label{fig:counterfactual_pipeline}
% \end{figure*}

To address the limitations of previous studies, we propose a novel multi-modality framework, \textbf{CAMEF}\footnote{\textit{The dataset and code for CAMEF are open-sourced at: https://github.com/lakebodhi/CAMEF}} (\textbf{c}ausal-\textbf{A}ugmented \textbf{M}ulti-Modality \textbf{E}vent-Driven Financial \textbf{F}orecasting). CAMEF integrates time-series and textual features through specially designed multi-feature fusion techniques, time-series decoding mechanisms, and causal learning strategies. By conducting a thorough review of financial literature, we identify six types of salient macroeconomic events for the forecasting analysis. Furthermore, the framework employs causal data augmentation powered by Large Language Models (LLMs) and a causal contrastive learning approach to enhance the causal understanding and forecast accuracy of CAMEF. This paper offers three key contributions:

\begin{itemize}

    \item \textbf{Novel Dataset:} We introduce a novel open-source synthetic dataset comprising 6 types of macroeconomic event scripts (ref to Tab. \ref{tab:data_summary} for details) from 2008 to April 2024 through reviewing from financial literature \cite{https://doi.org/10.1111/jofi.12818,GERTLER2018336,https://doi.org/10.1111/joes.12550,RePEc:ijc:ijcjou:y:2016:q:4:a:6,TADLE2022106021,RePEc:fip:fednep:00004,https://doi.org/10.1111/jofi.12196,ROSA2011915,NBERw1296,Gilbert2010-dt,GILBERT201778,RePEc:fip:fednci:y:2008:i:aug:n:v.14no.6,9fdfa12d09ae4792a11e8360a71a356a,RePEc:snb:snbwpa:2013-11}, alongside intra-day \textunderscore{high-frequency} financial data at 5-minute intervals from key U.S. stock indexes and Treasury bonds. To support causal learning, the dataset also includes counterfactual event scripts generated using our LLM-based  causal argumentation prompting, making it the first to integrate policy texts, high-frequency trading data, and causally augmented content.

    \item \textbf{Novel Multi-Modality Model:} We propose a novel multi-modality approach, CAMEF, that integrates time-series and textual features, incorporating specifically designed multi-feature fusion  and time-series decoding networks, which have been demonstrated to be effective for forecasting. Additionally, the model includes a causal learning mechanism to enhance forecasting capability by capturing the causal relationships between events and market reactions. 

    \item \textbf{Counterfactual Generation and Learning:} We introduce a counterfactual data augmentation strategy to generate counterfactual event scripts based on collected macroeconomic releases. This approach leverages LLMs to create scripts with varying sentiment levels by modifying key numerical values and sentiment-relevant phrases, while preserving the original format, writing style, and neutral words of the factual reports. Counterfactual events enable CAMEF to better understand the causal relationships between events and market reactions by learning from hypothetical scenarios, thereby improving its forecasting ability.

\end{itemize}

\section{Related Work}
\subsection{Event-Drive Financial Forecasting} 
Event-driven financial forecasting \cite{BAO2025102616} focuses on predicting asset prices \cite{RePEc:snb:snbwpa:2013-11,Gilbert2010-dt} and market volatility \cite{https://doi.org/10.1111/jofi.12196,https://doi.org/10.1111/jofi.12818} based on events like macroeconomic releases \cite{Gilbert2010-dt}, news \cite{liu2018leveragingfinancialnewsstock}, corporate announcements \cite{zhou-etal-2021-trade}, and social media activity \cite{xu-cohen-2018-stock}. Three main approaches exist in this area. The first leverages text analysis to predict asset responses based on event-related text. Early works utilized TF-IDF \cite{1196287,LI2014826} and topic models \cite{si-etal-2013-exploiting,NGUYEN20159603}, progressing to RNN-based models \cite{10.1145/3155133.3155202,liu2018leveragingfinancialnewsstock} and pre-trained transformers \cite{zhou-etal-2021-trade,shah-etal-2023-trillion}, which capture nuanced semantics. Although these models excel at semantic extraction, they often lack integration with historical price data, crucial for holistic forecasting.

The second line of approaches uses statistical and sequential models on numerical data, such as linear regression \cite{8212716}, ARIMA \cite{7046047}, and GARCH \cite{HYUPROH2007916}. Later, deep learning methods like RNNs \cite{liu2018leveragingfinancialnewsstock} and CNNs \cite{8126078,Durairaj2022} enhanced nonlinear modeling capabilities. More recently, transformer-based models, such as Informer \cite{Zhou_Zhang_Peng_Zhang_Li_Xiong_Zhang_2021} and FedFormer \cite{pmlr-v162-zhou22g}, improved long-range dependency modeling for time series data. However, these models tend to be ``case-specific,'' requiring task-specific training. In contrast, the lastest pre-trained models for time-series data, like MOMENT \cite{goswami2024moment}, Timer \cite{liu2024timer}, and TOKEN \cite{anonymous2024totem}, offer more generalized and adaptable solutions for time-series tasks.

The third line of research adopts multi-modality approaches, combining diverse data types to improve forecasting accuracy. Some studies incorporate text and audio \cite{qin-yang-2019-say,10.1145/3366423.3380128} but often overlook time-series dependencies. Recent work has integrated time-series and textual data; for example, \cite{10.1145/3394171.3413752,sawhney-etal-2020-deep} employed SVM and GRU models to capture time-series features. However, these models are relatively shallow for extracting complex patterns. More recent studies \cite{lee2024moat, Jia_Wang_Zheng_Cao_Liu_2024} leverage transformer-based models for time-series analysis, better capturing deeper temporal structures. Building on these advancements, this paper aims to utilize state-of-the-art pre-trained models with enhanced feature fusion and causal learning for multi-modality forecasting.

\subsection{Salient Macroeconomic Factors}

\textbf{Which macroeconomic announcements have a greater impact on financial markets than others?} This question has been widely studied in the financial literature, with Central Bank Communications standing out as the most-researched factor \cite{https://doi.org/10.1111/jofi.12818,GERTLER2018336,https://doi.org/10.1111/joes.12550,RePEc:ijc:ijcjou:y:2016:q:4:a:6,TADLE2022106021,RePEc:fip:fednep:00004,https://doi.org/10.1111/jofi.12196,ROSA2011915}. Beyond central bank communications, various other macroeconomic factors have also been identified as significant drivers of market movements. Among these, Non-farm Payrolls, Unemployment Releases, Initial Unemployment Claims, ISM Manufacturing Index, GDP Advance Releases, Consumer Confidence Index, and Producer Price Index (PPI) Reports have been found to notably influence price movements and market volatility through empirical statistical testings \cite{NBERw1296,Gilbert2010-dt,GILBERT201778,RePEc:fip:fednci:y:2008:i:aug:n:v.14no.6,9fdfa12d09ae4792a11e8360a71a356a,RePEc:snb:snbwpa:2013-11}. In this paper, we aim to leaverage the most significant factors evidented by the past financial literautre \cite{NBERw1296,Gilbert2010-dt,GILBERT201778,RePEc:fip:fednci:y:2008:i:aug:n:v.14no.6,9fdfa12d09ae4792a11e8360a71a356a,RePEc:snb:snbwpa:2013-11}, which include FOMC Meeting Documents, Non-farm Payrolls, Unemployment Releases, Initial Unemployment Claims, ISM Manufacturing Index, GDP Advance Releases, Consumer Confidence Index, and Producer Price Index (PPI) Reports.

\subsection{Counterfactual Data Augmentation by LLMs} 

Counterfactual Data Augmentation seeks to reduce spurious correlations and enhance model robustness. \citet{Kaushik2020Learning} introduced a method that augments training data with counterfactuals written by human annotators, effectively helping to mitigate spurious patterns. \citet{wu-etal-2021-polyjuice, ross-etal-2022-tailor} later proposed the use of hand-crafted templates and trained text generators to create counterfactual data through predefined perturbation types. However, these methods are limited by their reliance on fixed perturbations. More recently, \citet{chen-etal-2023-disco, wang-etal-2023-self-instruct} proposed more flexible, LLM-based approaches that leverage specifically designed in-context learning prompts and generation pipelines for counterfactual and instruction data generation. Following this direction, we present a counterfactual generation framework specifically designed for macroeconomic releases.

\begin{table*}[t]
\caption{Summary of Macroeconomic and Time-Series Data Types, Characteristics and Sources}
\label{tab:data_summary}
\begin{tabular}{@{}cc@{}}
\begin{minipage}{\columnwidth} % Adjusted width to equalize both tables
\centering
\caption*{A. Macroeconomic Event Summary}
\label{tab:data_summary_A}
\resizebox{\columnwidth}{!}{
\begin{tabular}{@{}lcccccl@{}}
\toprule
\textbf{Event Type}                    & \textbf{Data Type} & \textbf{Frequency} & \textbf{Period}      & \textbf{No. of Events} & \textbf{No. of C.F.s} & \textbf{Source}        \\ \midrule
\textbf{FOMC}                          & Html               & Quarterly          & 1993.3 $\sim$2024.6  & 255                    & 2,550                 & \makecell[l]{www.federal\\reserve.gov}  \\
\makecell[l]{\textbf{Unemployment} \\ \textbf{Insurance Claims}} & PDF, Txt & Weekly  & 2002.10 $\sim$2024.6 & 913 & 9,130 & oui.doleta.gov \\
\textbf{Employment Situation}          & Html, Txt          & Monthly            & 1994.2 $\sim$2024.6  & 363                    & 3,630                 & www.bls.gov            \\
\textbf{GDP Advance Report}            & Html               & Monthly            & 1996.8 $\sim$2024.6  & 333                    & 3,330                 & www.bea.gov            \\
\textbf{CPI Report}                    & Html, Txt          & Monthly            & 1994.2 $\sim$2024.6  & 357                    & 3,570                 & www.bls.gov            \\
\textbf{PPI Report}                    & Html, Txt          & Monthly            & 1994.2 $\sim$2024.6  & 348                    & 3,480                 & www.bls.gov            \\ \bottomrule
\end{tabular}
}
\end{minipage}

& % Separator for the two tables

\begin{minipage}{\columnwidth} % Same width for consistency
\centering
\caption*{B. Time-Series Data Summary}
\label{tab:data_summary_B}
\resizebox{\columnwidth}{!}{
\begin{tabular}{@{}lcccc@{}}
\toprule
\textbf{Time Series Data}                & \textbf{Data Types}    & \textbf{Frequency} & \textbf{Range}        & \textbf{No. of Data Points} \\ \midrule
\textbf{SP500 (SPX)}             & Open, Close, High, Low & 5 Min              & 2008.01 $\sim$2024.06 & 331,257                     \\
\textbf{Dow Industrial (INDU)}    & Open, Close, High, Low & 5 Min              & 2012.07 $\sim$2024.06 & 263,445                     \\
\textbf{NASDAQ (NDX)}    & Open, Close, High, Low & 5 Min              & 2008.01 $\sim$2024.06 & 332,616                     \\
\makecell[l]{\textbf{US Treasury Bond} \\ \textbf{at 1-Month (USGG1M)}}    & Open, Close, High, Low & 5 Min              & 2013.01 $\sim$2024.06 & 751,443                     \\
\makecell[l]{\textbf{US Treasury Bond} \\ \textbf{at 5-Year (USGG5YR)}}    & Open, Close, High, Low & 5 Min              & 2013.01 $\sim$2024.06 & 734,773                     \\
\bottomrule
\end{tabular}
}
\end{minipage}
\end{tabular}
\end{table*}

\section{Problem Formulation}
\label{sec:definition}
\indent \textbf{Event Set} is defined as $\mathbb{E} := \{ \mathcal{E}_{1}, \mathcal{E}_{2}, \dots, \mathcal{E}_{|\mathbb{E}}| \} $, where $\mathbb{E}$ represents a collection of $|\mathbb{E}|$ event scripts. Each event $\mathcal{E}_{i}$ occurs at a specific timestamp $i$ and belongs to one of the event types shown in Table \ref{tab:data_summary}A. Each event script $\mathcal{E}_{i}$ consists of a sequence of word tokens, represented as $\mathcal{E}_{i} := \{w_1, w_2, \dots, w_m \}$.

\textbf{Time Series Data} is defined as $\mathcal{X} := \{ X_1, X_2, \dots, X_{|\mathcal{X}|} \}$, where each $X_i$ represents the numerical data at time step $i$. An event $\mathcal{E}_i$ is \textbf{aligned} with a time series segment $\mathcal{X}_{i-\tau:i+\tau}$, where $i$ denotes the time of the releasement of the event, and $\tau$ represents the duration of the time-series segment both preceding and succeeding time step $i$, denoted as \([\mathcal{X}_{i-\tau:i+\tau} \mapsto \mathcal{E}_i ]\). This alignment reflects the time series segment leading up to the event ($\mathcal{X}_{i-\tau}$) and the period during which the event is expected to have an effect ($\mathcal{X}_{i+\tau}$). 

\textbf{Event-Driven Forecasting}: Given a dataset \( \mathcal{U} = \{ [\mathcal{X}_{i-\tau:i+\tau} \mapsto \mathcal{E}_i] \}_{i=1}^n \) consisting of \( n \) aligned event and time-series pairs, for each data pair in \( \mathcal{U}\), the model uses both the event text \( \mathcal{E}_i \) and the historical time-series segment \( \mathcal{X}_{i-\tau:i} \), which spans \( \tau \) steps before the event’s release at time \( i \), to forecast the future time steps \( \mathcal{X}_{i+1:i+\tau} \).

\begin{figure}[t]
    \centering
    % Subfigure 1
    \begin{subfigure}{0.35\columnwidth}
        \centering
        \includegraphics[width=\linewidth]{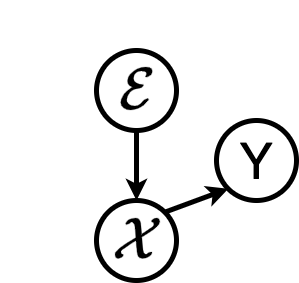}
        \caption{\centering Causal Effect Graph}
        \label{fig:scm}
    \end{subfigure}
    \hfill % Adds spacing between subfigures
    % Subfigure 2
    \begin{subfigure}{0.5\columnwidth}
        \centering
        \includegraphics[width=\linewidth]{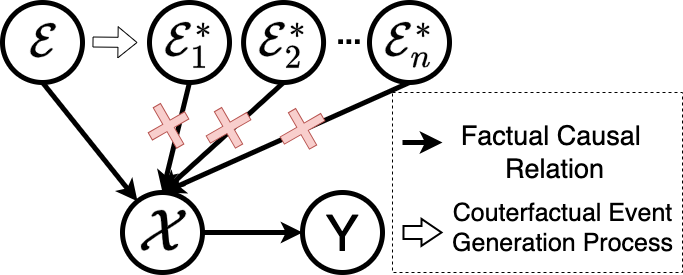}
        \caption{\centering Cause-Effect Detection with Counterfactual Events}
        \label{fig:CE}
    \end{subfigure}
    % Main Figure Caption
    \caption{The illustration of causal relationships and counterfactual events.}
    \label{fig:main}
\end{figure}

\textbf{Causal Effect Graph:} A Causal Effect Graph represents the causal links among variables: textual modality \(\mathcal{E}\) (event scripts), current price trend \(\mathcal{X}\), and future time series movements \(\mathbf{Y}\). In event-driven financial prediction, events influence the market movements of financial assets, forming a causal chain denoted as \(\mathcal{E} \rightarrow \mathcal{X} \rightarrow \mathbf{Y}\), as illustrated in Figure \ref{fig:scm}.

\textbf{Counterfactual Event:} A Counterfactual Event (CE) represents a modified event script in which key variables (e.g., unemployment rates, GDP values) are altered relative to the factual event, while the surrounding context remains unchanged. These events are denoted as \(\{\mathcal{E}^{*}_{1}, \mathcal{E}^{*}_{2}, \ldots, \mathcal{E}^{*}_{n}\}\). CEs are utilized to train CAMEF, enabling it to identify factual cause-effect relationships, represented as \(\mathcal{E} \rightarrow \mathcal{X} \rightarrow \mathcal{Y}\), as illustrated in Figure \ref{fig:CE}.

% \section{Data Collection and Counterfactual Event Augmentation}
% \label{sec:data_acquisition}
% In this section, we introduce the proposed dataset and the counterfactual augmentation methodology. Following the findings in the financial literature \cite{https://doi.org/10.1111/jofi.12818,GERTLER2018336,https://doi.org/10.1111/joes.12550,RePEc:ijc:ijcjou:y:2016:q:4:a:6,TADLE2022106021,RePEc:fip:fednep:00004,https://doi.org/10.1111/jofi.12196,ROSA2011915}, we selected 6 most significant macroeconomic announcements, including the \textbf{\textit{FOMC Minutes, Unemployment Insurance Releases, Initial Unemployment Claims, GDP Advance Releases, CPI Report, and PPI Report}}. Whereas for the time-series dataset,  we focused on 5 key financial assets, including major U.S. stock indexes and Treasury bonds, which are mostly adopted from the macroeconmic researchs \cite{https://doi.org/10.1111/jofi.12818,GERTLER2018336,https://doi.org/10.1111/joes.12550,RePEc:ijc:ijcjou:y:2016:q:4:a:6,TADLE2022106021,RePEc:fip:fednep:00004,https://doi.org/10.1111/jofi.12196,ROSA2011915}, including , as outlined in Table \ref{tab:data_summary_B}. 

\section{Data Collection and Counterfactual Event Augmentation}
\label{sec:data_acquisition}

This section introduces the proposed dataset and the methodology for counterfactual event augmentation. The dataset includes 6 types of key macroeconomic announcements ranging from 2004 to 2024, selected through an extensive review of the financial literature, along with \textunderscore{high-frequency} trading data. Unlike the daily-based trading data used in previous studies, this high-frequency data provides more predictive accuracy and better reflects real trading behavior in the industry.

\subsection{Dataset Acquisition}

The primary question guiding the collection of this dataset is: \textbf{``Which macroeconomic releases have the greatest impact on financial markets?''} To address this, we conducted a comprehensive review of the financial literature to identify key macroeconomic factors that influence market behavior. Several dominant factors emerged, including the \textbf{FOMC Minutes} \cite{https://doi.org/10.1111/jofi.12818, GERTLER2018336, https://doi.org/10.1111/joes.12550, RePEc:ijc:ijcjou:y:2016:q:4:a:6, TADLE2022106021, RePEc:fip:fednep:00004, https://doi.org/10.1111/jofi.12196, ROSA2011915}, along with \textbf{Unemployment Insurance Claims}, \textbf{Employment Situation Reports}, \textbf{GDP Advance Releases}, and the \textbf{Consumer Price Index (CPI)} and \textbf{Producer Price Index (PPI)} reports \cite{NBERw1296, Gilbert2010-dt, GILBERT201778, RePEc:fip:fednci:y:2008:i:aug:n:v.14no.6, 9fdfa12d09ae4792a11e8360a71a356a, RePEc:snb:snbwpa:2013-11}, which serve as the textual modality data for our dataset. 

To collect these data, we developed web crawlers to extract raw files directly from official sources, including HTML, PDF, and TXT formats. These raw files were then pre-processed and converted into a structured and unified text format, ensuring consistency and ease of subsequent analysis. Table \ref{tab:data_summary_A} provides a summary of the data types, collection frequencies, time periods, and sources of the events included in our dataset. Further details on the data crawling and pre-processing methodologies can be found in Appendix B.

In addition to the textual data, studies \cite{https://doi.org/10.1111/jofi.12818, GERTLER2018336, https://doi.org/10.1111/joes.12550, RePEc:ijc:ijcjou:y:2016:q:4:a:6, TADLE2022106021, RePEc:fip:fednep:00004, https://doi.org/10.1111/jofi.12196, ROSA2011915} have demonstrated that the largest market impacts are typically observed in major U.S. stock indexes and Treasury bonds. Therefore, we focused on collecting high-frequency trading time-series data at 5 minute interval for key stock indexes, including the \textbf{S\&P 500 (SPX)}, \textbf{Dow Industrial (INDU)}, \textbf{NASDAQ (NDX)}, as well as \textbf{U.S. Treasury Bond at 1-Month (USGG1M)} and \textbf{Treasury Bond at 5-Year (USGG5YR)}.

\begin{figure*}[ht]
    \centering
    \includegraphics[width=\linewidth]{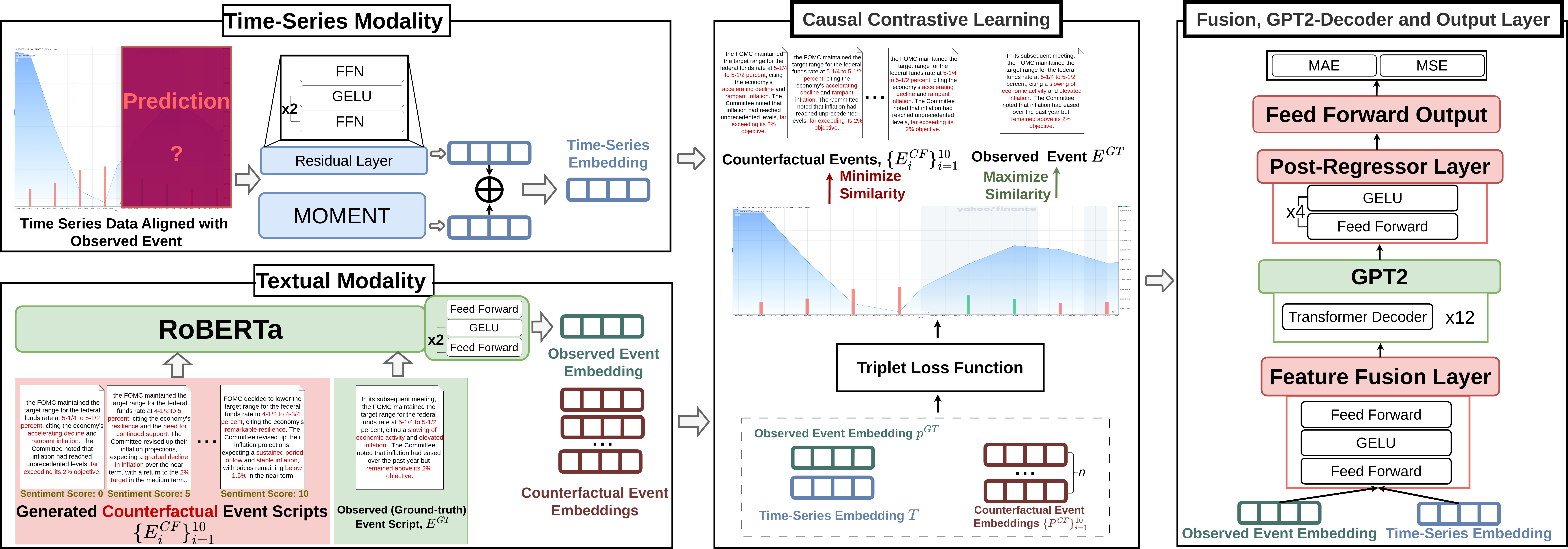}
    \caption{The Pipeline and Neural Architecture of CAMEF}
    \label{fig:model_architecture}
\end{figure*}

% We expect the modifications to fulfill the following requirements:

% \begin{itemize}
%     \item \textbf{Numerical Reasonability:} Adjust key numerical variables logically to reflect the target sentiment \( S_i^\prime \), such as higher unemployment rates for negative sentiment or lower rates for positive sentiment, while maintaining factual plausibility.
%     \item \textbf{Sentiment Relevance:} Modify sentiment-relevant words and phrases (\( \mathcal{E}_i^\text{sentiment} \)) to align with the target sentiment while ensuring contextual coherence and consistency.
%     \item \textbf{Structural Consistency:} Preserve the original script's format, style, and structure to ensure plausibility, readability, and coherence of the counterfactual event.
% \end{itemize}

\subsection{Counterfactual Events Generation based on LLM}
\label{sec:CE_generation}
This section describes the process of counterfactual event generation, creating  hypothetical scenarios from existing event scripts. The aim is to reflect a target sentiment of a given event script while maintaining logical consistency and coherence of the original script. Our goal is modifying sentiment-relevant elements (such as key facts, sentiment-indicative phrases, or numerical values) without disrupting the sentiment-neutral components of the script.

Formally, for a given event script \( \mathcal{E}_i := \{w_1, w_2, \dots, w_m\} \), the objective is to produce a counterfactual version \( \mathcal{E}_i^\prime \) that embodies the desired target sentiment \( S_i^\prime \). Conceptually, the event script can be viewed as comprising sentiment-relevant content (\( \mathcal{E}_i^\text{sentiment} \)) and sentiment-neutral content (\( \mathcal{E}_i^\text{neutral} \)), so that \( \mathcal{E}_i = \mathcal{E}_i^\text{neutral} \cup \mathcal{E}_i^\text{sentiment} \). Instead of explicitly decomposing the script, we guide a language model (LLM) using structured prompts to modify only the sentiment-relevant content. This ensures that neutral content remains intact or is replaced with semantically equivalent expressions. Formally:
\begin{equation}
\mathcal{E}_i^\prime = \mathcal{E}_i^\text{neutral} \cup f_\text{LLM}(\mathcal{E}_i^\text{sentiment} \mid S_i^\prime),
\end{equation}

\noindent where \( f_\text{LLM} \) adjusts \( \mathcal{E}_i^\text{sentiment} \) to align with \( S_i^\prime \), and \( \mathcal{E}_i^\text{neutral} \) remains unchanged or is replaced by equivalent expressions.

Specifically, we used the LLaMA-3 8B model with a series of carefully designed prompts. These prompts include three key steps, with detailed templates provided in Appendix A:

\begin{enumerate}
    \item \textbf{Summarization Prompt (Appendix A.1):} Condenses lengthy event scripts into concise summaries, addressing memory constraints while retaining sentiment-relevant content and key numerical variables.
    \item \textbf{Sentiment Analysis Prompt (Appendix A.2):} Assigns a sentiment score (from 1, very negative, to 10, very positive) to the original event script. This score provides a baseline for generating counterfactual versions.
    \item \textbf{Counterfactual Generation Prompt (Appendix A.3):} Produces multiple counterfactual scripts, each reflecting a different sentiment level. The prompt modifies sentiment-related phrases and numerical values (\( \mathcal{E}_i^\text{sentiment} \)) while preserving or equivalently substituting neutral content (\( \mathcal{E}_i^\text{neutral} \)). This approach ensures numerical reasonability, sentiment relevance, and structural consistency.
\end{enumerate}

This multi-step prompt strategy facilitates the generation of coherent, contextually relevant counterfactual events, enabling exploration of diverse market scenarios and deeper causal understanding.

\section{CAMEF Architecture}
\label{sec:model_architecture}
The CAMEF model integrates both textual and time-series information through a structured architecture consisting of a textual encoder, a time-series encoder, and a forecasting decoder, as depicted in Fig. \ref{fig:model_architecture}. Each component is detailed below.

\subsection{Textual Modality Encoder (\( \textbf{CAMEF}_{\text{Textual}}\))}
\label{sec:textual_encoder}
We encode event scripts using RoBERTa \cite{liu2020roberta}. Given an input script $\mathbf{E}_i = {w_1, w_2, \dots, w_m}$, RoBERTa produces contextual token embeddings:

\begin{equation} 
\{ \mathbf{h}_1, \mathbf{h}_2, \dots, \mathbf{h}_m \} = \text{RoBERTa}(\{w_1, w_2, \dots, w_m\}),
\end{equation}

\noindent where $\mathbf{h}_j \in \mathbb{R}^{1 \times 768}$ is the embedding of token $w_j$. Each embedding is passed through a projection network with three linear layers and GELU activations:

\begin{equation} 
\mathbf{e}_j = \mathbf{W}^{(3)} \cdot \text{GELU}(\mathbf{W}^{(2)} \cdot \text{GELU}(\mathbf{W}^{(1)} \mathbf{h}_j + \mathbf{b}^{(1)}) + \mathbf{b}^{(2)}) + \mathbf{b}^{(3)},
\end{equation}

\noindent where $\mathbf{W}^{(1)} \in \mathbb{R}^{768 \times 1024}$, $\mathbf{W}^{(2)} \in \mathbb{R}^{1024 \times 1024}$, and $\mathbf{W}^{(3)} \in \mathbb{R}^{1024 \times 768}$. The final encoding vector $\mathbf{E}_i$ for the script is computed as the average of the transformed embeddings, 
$\mathbf{E}_i = \frac{1}{m} \sum_{j=1}^m \mathbf{e}_j$.

% This averaging step condenses the sequence of embeddings into a single vector, providing a summarized representation of the event script.

% \subsection{Textual Modality Encoder}

% The event scripts are transformed into vector representations using the RoBERTa model \cite{liu2020roberta}, which is an efficient textual information encoder based on transformer architecture. Given an input event script from the event collection defined in Sec. \ref{sec:definition}, denoted as $\mathbf{E}_i := {w_1, w_2, \dots, w_m}$, the tokens are encoded into corresponding vector representations. The output from RoBERTa provides a sequence of token embeddings:

% \begin{equation}
% \{\mathbf{w}_1, \mathbf{w}_2, \dots, \mathbf{w}_m\} = \textbf{RoBERTa}(\{w_1, w_2, \dots, w_m\}). 
% \end{equation}
% \noindent To generate the final encoding vector for the event script, we apply average pooling over the sequence of token embeddings:

% \begin{equation}
%     \mathbf{E}_i = \frac{1}{m} \sum_{j=1}^{m} \mathbf{w}_j,
% \end{equation}

% \noindent where, $\mathbf{\mathcal{E}}_i$ represents the final pooled vector that summarizes the event script $\mathbf{E}_i$, where $m$ is the number of tokens in the script.

\subsection{Time-Series Modality Encoder \\ ( \( \textbf{CAMEF}_{\text{Time-Series}} \) )} 
\label{sec:time-series_encoder}
To encode the time series data, we employ a pretrained time series encoder, MOMENT \cite{goswami2024moment}, which generates a fixed-dimensional vector for an input time series segment. Subsequently, we design a multi-residual layer to further refine the encoding vectors, as shown below:

\begin{equation}
    \mathbf{\mathcal{X}}_i = \textbf{MOMENT}(\{ X_1, X_2, \dots, X_n\}),
\end{equation}

\noindent where $\mathbf{\mathcal{X}}_i \in \mathbb{R}^{d}$ is the encoded vector for the input time series segment $\{ X_1, X_2, \dots, X_n\}$, and \(d\) represents the dimensionality of the encoded vector. To enhance this representation, we introduce a multi-residual projection layer:

\begin{equation}
    \mathbf{Z}_i = \mathbf{\mathcal{X}}_i + f_{\text{residual}}(\mathbf{\mathcal{X}}_i),
\end{equation}

\noindent where \(f_{\text{residual}}(\mathbf{\mathcal{X}}_i)\) represents the transformation applied through the residual projection layer, which consists of multiple linear layers interleaved with GELU activations:

\begin{equation}
    f_{\text{residual}}(\mathbf{\mathcal{X}}_i) = \mathbf{W}_3 \cdot \text{GELU}(\mathbf{W}_2 \cdot \text{GELU}(\mathbf{W}_1 \cdot \mathbf{\mathcal{X}}_i + \mathbf{b}_1) + \mathbf{b}_2) + \mathbf{b}_3,
\end{equation}

\noindent where \(\mathbf{W}_1, \mathbf{W}_2 \in  \mathbb{R}^{1024 \times 1024}, \mathbf{W}_3 \in \mathbb{R}^{1024 \times 768}\) are the weight matrices, and \(\mathbf{b}_1, \mathbf{b}_2  \in \mathbb{R}^{1024}, \mathbf{b}_3 \in \mathbb{R}^{768}\) are the respective biases. Finally, \(\mathbf{Z}_i \in \mathbb{R}^{1024}\) serves as the refined vector for the time series segment.

\subsection{Feature Fusion and Time Series Decoder}

After obtaining the encoded vectors from the textual and time series data, denoted as $\mathbf{E}_i$ and $\mathbf{Z}_i $, respectively, we concatenate them to form a unified representation:

\begin{equation}
\mathbf{E}_{\text{combined}} = \text{Concat}(\mathbf{E}_i, \mathbf{Z}_i),
\end{equation}

\noindent which captures both semantic content from macroeconomic texts and temporal patterns from time series inputs. To fuse these modalities, $\mathbf{E}_{\text{combined}}$ is passed through a two-layer feedforward network with GELU activation:

\begin{equation}
\mathbf{E}_{\text{fused}} = \mathbf{W}^{(f2)} \cdot \text{GELU}(\mathbf{W}^{(f1)} \cdot \mathbf{E}*{\text{combined}} + \mathbf{b}^{(f1)}) + \mathbf{b}^{(f2)},
\end{equation}

\noindent where $\mathbf{W}^{(f1)} \in \mathbb{R}^{(2 \times 768) \times 1024}$, $\mathbf{W}^{(f2)} \in \mathbb{R}^{1024 \times 1024}$, and $\mathbf{b}^{(f1)}, \mathbf{b}^{(f2)} \in \mathbb{R}^{1024}$ are the corresponding weight matrices and bias terms. This fusion block enables interaction across modalities and produces a refined joint embedding for downstream decoding.

We then employ GPT-2 \cite{radford2019language} as the decoder to decode the fused vector by leveraging its effective auto-regressive ability:

\begin{equation}
\mathbf{H}^{(l)} = f_{\text{GPT2\_layer}}^{(l)}(\mathbf{H}^{(l-1)}),
\end{equation}

\noindent where \(\mathbf{H}^{(0)} = \mathbf{E}_{\text{fused}}\), and \(f_{\text{GPT2\_layer}}^{(l)}\) represents the transformation function of the \(l\)-th GPT-2 layer, where \( l=12\). After the final layer, the output is normalized using a layer normalization function:

\begin{equation}
\label{eqn:output_embedding}
\mathbf{H}_{\text{final}} = \text{LayerNorm}(\mathbf{H}^{(l)}),
\end{equation}

\noindent The final output \(\mathbf{H}_{\text{final}}\) is then used to generate predictions based on the combined multi-modal information.

\subsection{Time-Series Forecasting Post-Regressor and Learning Objectives}

We designed a \textbf{Post-Regressor} that applies a linear transformation to the concatenated vector \(\mathbf{H}_{\text{final}}\), followed by GELU activation and a dropout layer with a rate of 0.1:

\begin{equation}
\mathbf{R}^{(k)} = \text{GELU}(\mathbf{W}^{(k)} \cdot \mathbf{R}^{(k-1)} + \mathbf{b}^{(k)}),
\end{equation}

\noindent where \(\mathbf{R}^{(0)} = \mathbf{H}_{\text{final}}\), \(\mathbf{W}^{(k)}\) and \(\mathbf{b}^{(k)}\) are the weight matrix and bias of the \(k\)-th linear layer, respectively. \(k = 4 \) is the total number of layers in the regressor. The final linear layer maps the representation to a vector of shape \((d \times \text{pred\_len})\), where \(d\) is the forecast dimensionality and \(\text{pred\_len}\) is the number of predicted time steps:

\begin{equation}
\mathbf{\hat{Y}} = \mathbf{W}_{\text{out}} \cdot \mathbf{R}^{(K)} + \mathbf{b}_{\text{out}},
\end{equation}

\noindent where \(\mathbf{\hat{Y}}\), represents the predicted time series values.

\textbf{Learning Objectives for Time Series:} We employ a combination of Mean Squared Error (MSE) loss and Mean Absolute Error (MAE) loss to optimize the model. The MSE loss minimizes the squared differences between the predicted time series values, \(\mathbf{\hat{Y}}\), and the ground truth values, \(\mathbf{Y}\), while the MAE loss minimizes the absolute differences. These are defined as:

\begin{equation}
\mathcal{L}_{\text{MSE}} = \frac{1}{n} \sum_{i=1}^{n} (\mathbf{\hat{Y}}_i - \mathbf{Y}_i)^2, \quad
\mathcal{L}_{\text{MAE}} = \frac{1}{n} \sum_{i=1}^{n} |\mathbf{\hat{Y}}_i - \mathbf{Y}_i|,
\end{equation}

\noindent where \(n\) is the number of predicted values (e.g., 35, 70, or 140, as defined in Section \ref{sec:exp_setting}). The total loss function combines both objectives to balance optimization for large and small errors:

\begin{equation}
\label{eqn:time_objective}
\mathcal{L}_{\text{Time}} = \mathcal{L}_{\text{MSE}} + \mathcal{L}_{\text{MAE}}.
\end{equation}

\begin{table*}[ht]
\caption{Financial Forecasting Results (MSE and MAE Scores) for CAMEF and Baselines Across Various Financial Assets: S\&P500 (SPX), Dow Industrials (INDU), Nasdaq100 (NDX) Index, US 1-Month Treasury Bond (USGG1M), and US 5-Year Treasury Bond (USGG5YR).}
\label{tab:main_results}
\resizebox{\textwidth}{!}{%
\begin{tabular}{c|c|cl|cl|cl|cl|cl}
\hline
\multicolumn{1}{l|}{\multirow{2}{*}{\textbf{Model / Datasets}}} & \multirow{2}{*}{\textbf{\begin{tabular}[c]{@{}c@{}}Forecasting\\ Length\end{tabular}}} & \multicolumn{2}{c|}{\textbf{SP500 (SPX)}} & \multicolumn{2}{c|}{\textbf{Dow Industrial (INDU)}} & \multicolumn{2}{c|}{\textbf{NASDAQ (NDX)}} & \multicolumn{2}{c|}{\textbf{USGG1M}} & \multicolumn{2}{c}{\textbf{USGG5YR}} \\ \cline{3-12} 
\multicolumn{1}{l|}{} &  & MSE & \multicolumn{1}{c|}{MAE} & MSE & \multicolumn{1}{c|}{MAE} & MSE & \multicolumn{1}{c|}{MAE} & MSE & \multicolumn{1}{c|}{MAE} & MSE & \multicolumn{1}{c}{MAE} \\ \hline
\multirow{3}{*}{\textbf{ARIMA}} & 35 & 0.0032628 & 0.0308016 & 0.0121513 & 0.0523810 & {\ul 0.0065253} & 0.0435694 & 0.0072223 & 0.0262512 & 0.0038973 & 0.0340227 \\
 & 70 & 0.0035361 & 0.0352004 & 0.0139245 & 0.0656002 & {\ul 0.0082324} & 0.0520933 & 0.0028710 & 0.0304132 & 0.0039441 & 0.0366630 \\
 & 140 & 0.0051080 & 0.0439935 & 0.0219147 & 0.0793471 & {\ul 0.0118692} & 0.0665155 & 0.0089949 & 0.0338275 & 0.0050746 & 0.0455617 \\ \hline
\multirow{3}{*}{\textbf{DLinear}} & 35 & 0.0144331 & 0.0896910 & 0.0395136 & 0.1380539 & 0.0183989 & 0.0999178 & 0.0108576 & 0.0706170 & 0.0147211 & 0.0876379 \\
 & 70 & 0.0120578 & 0.0817373 & 0.0406573 & 0.1282699 & 0.0189431 & 0.0972439 & 0.0093591 & 0.0678380 & 0.0146430 & 0.0881574 \\
 & 140 & 0.0178138 & 0.0931039 & 0.0747210 & 0.1724027 & 0.0344153 & 0.1287803 & 0.0101465 & 0.0645396 & 0.0179185 & 0.0977673 \\ \hline
\multirow{3}{*}{\textbf{Autoformer}} & 35 & 0.0068136 & 0.0540556 & 0.0277636 & 0.0975948 & 0.0135249 & 0.0796486 & 0.0047505 & 0.0388076 & 0.0092717 & 0.0622862 \\
 & 70 & 0.0088997 & 0.0628341 & 0.0375279 & 0.1185710 & 0.0185264 & 0.0933398 & 0.0065554 & 0.0471531 & 0.0113750 & 0.0727373 \\
 & 140 & 0.0158248 & 0.0829188 & 0.0772580 & 0.1640365 & 0.0334504 & 0.1262163 & 0.0086212 & 0.0533337 & 0.0152298 & 0.0875821 \\ \hline
\multirow{3}{*}{\textbf{FEDformer}} & 35 & 0.0072377 & 0.0576221 & 0.0304808 & 0.1044447 & 0.0128668 & 0.0758742 & 0.0063596 & 0.0519141 & 0.0094995 & 0.0494306 \\
 & 70 & 0.0088056 & 0.0621386 & 0.0399824 & 0.1229772 & 0.0171008 & 0.0889995 & 0.0062841 & 0.0448822 & 0.0099615 & 0.0664197 \\
 & 140 & 0.0157429 & 0.0819469 & 0.0786426 & 0.1677705 & 0.0313433 & 0.1214097 & 0.0083784 & 0.0518365 & 0.0131231 & 0.0782861 \\ \hline
\multirow{3}{*}{\textbf{iTransformer}} & 35 & 0.0064209 & 0.0516341 & 0.0270860 & 0.0927605 & 0.0125008 & 0.0751656 & 0.0011000 & 0.0155975 & \multicolumn{1}{l}{0.0056660} & \textbf{0.0183524} \\
 & 70 & 0.0069612 & 0.0540920 & 0.0304566 & 0.1038424 & 0.0151214 & 0.0816262 & 0.0021811 & 0.0221315 & \textbf{0.0011721} & \textbf{0.0226975} \\
 & 140 & 0.0128021 & 0.0718771 & 0.0680991 & 0.1486782 & 0.0254479 & 0.1047119 & 0.0052255 & 0.0327429 & \textbf{0.0017441} & \textbf{0.0282537} \\ \hline
\multirow{3}{*}{\textbf{PatchTST}} & 35 & 0.0063304 & 0.0507462 & 0.0293131 & 0.0989455 & 0.0122679 & 0.0764552 & 0.0012060 & 0.0163610 & 0.0063078 & 0.0520734 \\
 & 70 & 0.0072471 & 0.0547738 & \multicolumn{1}{l}{0.0339444} & 0.1116023 & 0.0153753 & 0.0824677 & 0.0021643 & 0.0223036 & 0.0079617 & 0.0606007 \\
 & 140 & 0.0130219 & 0.0712171 & 0.0688582 & 0.1452592 & 0.0256001 & 0.1047749 & 0.0054544 & 0.0341517 & 0.0118553 & 0.0747537 \\ \hline
\multirow{3}{*}{\textbf{GPT4MTS}} & 35 & \multicolumn{1}{l}{0.00795088} & 0.0674255 & {\ul 0.0026558} & 0.0417553 & 0.0011035 & 0.0240469 & 0.0017016 & 0.0309320 & 0.0028713 & 0.0389277 \\
 & 70 & \multicolumn{1}{l}{0.00171038} & 0.0305950 & {\ul 0.0027033} & {\ul 0.0393112} & 0.0016205 & 0.0336116 & 0.0019098 & 0.0279222 & 0.0023371 & 0.0354777 \\
 & 140 & \multicolumn{1}{l}{0.00212612} & {\ul 0.0330497} & {\ul 0.0045029} & {\ul 0.0458267} & 0.0025175 & 0.0341671 & 0.0013648 & 0.0260887 & 0.0037004 & 0.0449272 \\ \hline
\multirow{3}{*}{\textbf{TEST}} & 35 & \multicolumn{1}{l}{{\ul 0.00073333}} & {\ul 0.0199733} & \multicolumn{1}{l}{0.0026572} & \textbf{0.0296887} & 0.0006593 & {\ul 0.0194091} & {\ul 0.0003252} & {\ul 0.0137511} & {\ul 0.0013633} & 0.0265036 \\
 & 70 & \multicolumn{1}{l}{{\ul 0.00078762}} & {\ul 0.0205412} & 0.0088903 & 0.0599179 & 0.0010678 & {\ul 0.0248594} & {\ul 0.0010515} & {\ul 0.0182706} & 0.0034630 & 0.0415002 \\
 & 140 & \multicolumn{1}{l}{{\ul 0.00278572}} & 0.0467150 & 0.0070749 & 0.0557744 & 0.0020130 & {\ul 0.0362325} & {\ul 0.0006995} & {\ul 0.0207850} & 0.0024356 & 0.0375164 \\ \hline
\multirow{3}{*}{\textbf{CAMEF}} & 35 & \multicolumn{1}{l}{\textbf{0.00048860}} & \textbf{0.0154050} & \multicolumn{1}{l}{\textbf{0.0025349}} & {\ul 0.0366245} & \multicolumn{1}{l}{\textbf{0.0005468}} & \textbf{0.0178845} & \multicolumn{1}{l}{\textbf{0.0002883}} & \textbf{0.0118010} & \multicolumn{1}{l}{\textbf{0.0013234}} & {\ul 0.0260618} \\
 & 70 & \multicolumn{1}{l}{\textbf{0.00064780}} & \textbf{0.0178691} & \multicolumn{1}{l}{\textbf{0.0025042}} & \textbf{0.0365500} & \multicolumn{1}{l}{\textbf{0.0005814}} & \textbf{0.0162882} & \multicolumn{1}{l}{\textbf{0.0004402}} & \textbf{0.0139699} & \multicolumn{1}{l}{{\ul 0.0020701}} & {\ul 0.0326371} \\
 & 140 & \multicolumn{1}{l}{\textbf{0.0010756}} & \textbf{0.0210284} & \multicolumn{1}{l}{\textbf{0.0039313}} & \textbf{0.0383459} & \multicolumn{1}{l}{\textbf{0.0010159}} & \textbf{0.0207716} & \multicolumn{1}{l}{\textbf{0.0004938}} & \textbf{0.0148485} & \multicolumn{1}{l}{{\ul 0.0022458}} & {\ul 0.0336680} \\ \hline
\end{tabular}
}
\end{table*}

\subsection{Counterfactual Events Sampling and Causal Learning Objective}

Causal learning enhances the robustness of the CAMEF model by enabling it to identify the correct event script among sampled counterfactual events (CEs). To achieve this, we first design a \textbf{Diverse Counterfactual Event Sampling Mechanism}, which generates two types of CEs. These counterfactuals, along with their corresponding time-series data, are then encoded using the textual and time-series modalities of CAMEF. This process helps the model learn causal relationships between events and their corresponding time-series movements.

\subsubsection{Diverse Counterfactual Event Sampling Mechanism}
We propose a \textbf{Diverse Counterfactual Event Sampling Mechanism} to enhance the model’s ability to both identify the ground-truth event and distinguish between different event types. This mechanism is designed with two objectives: (1) to help the model recognize the ground-truth event among similar counterfactuals of the same type, and (2) to enable the model to differentiate between events of different types. 

To achieve these objectives, we generate two categories of counterfactual events for each factual event:
\begin{enumerate}
    \item \textbf{Identical Type Sampling:} Counterfactual events of the same type as the ground-truth event, created by modifying sentiment-relevant components and key numerical variables, as detailed in Sec.~\ref{sec:CE_generation}.
    \item \textbf{Diverse Type Sampling:} Counterfactual events of a different type, sampled by substituting the ground-truth event with 5 other event type occurring on the closest date.
\end{enumerate}

This mechanism provides a diverse set of counterfactual events, collectively denoted as \(\mathbb{E}^{CF} := \{\mathcal{E}^{CF}_1, \mathcal{E}^{CF}_2, \ldots, \mathcal{E}^{CF}_{|\mathbb{E}^{CF}|}\}\), we set the total number of diverse-type samples to be 5, and the default number of identical-type samples to be 10 as introduced Sec.~\ref{sec:CE_generation}. 

\subsubsection{Causal Learning Objective}

The causal learning process utilizes both the textual (see Sec. \ref{sec:textual_encoder}) and time-series (see Sec. \ref{sec:time-series_encoder}) encoders of CAMEF to capture the relationships between events and market movements. The textual encoder is used to encode both the ground-truth event \(\mathcal{E}^{GT}\) and the sampled CEs \(\mathcal{E}^{CF}\):

\begin{equation}
\{\mathbf{P}^{GT}, \mathbf{P}^{CF}_1, \dots, \mathbf{P}^{CF}_{|\mathbb{E}^{CF}|}\} = \textbf{CAMEF}_{\text{Textual}}(\{\mathcal{E}^{GT}\} \cup \{\mathcal{E}^{CF}_i\}_{i=1}^{|\mathbb{E}^{CF}|}),
\end{equation}

\noindent where \(\mathbf{P}^{GT}\) represents the embedding of the ground-truth event, and \(\{\mathbf{P}^{CF}_i\}_{i=1}^{|\mathbb{E}^{CF}|}\) represents the embeddings of the sampled counterfactual events.

The time-series encoder is used to encode the historical time-series segment \(\mathcal{X}\) aligned with the ground-truth event, resulting in the time-series embedding:

\begin{equation}
\mathbf{T} = \textbf{CAMEF}_{\text{Time-Series}}(\mathcal{X}).
\end{equation}

\noindent \textbf{Triplet Loss:} The triplet loss is applied to enforce that the ground-truth event embedding \(\mathbf{P}^{GT}\) is closer to the time-series embedding \(\mathbf{T}\) than any counterfactual event embedding \(\mathbf{P}^{CF}_i\), by a margin \(\alpha\) (set to 1.0):

\begin{equation}
\mathcal{L}_{\text{Causal-TL}} = \max\big(0, d(\mathbf{P}^{GT}, \mathbf{T}) - d(\mathbf{P}^{CF}_i, \mathbf{T}) + \alpha\big),
\end{equation}

\noindent where \(d(\cdot, \cdot)\) denotes the distance between two embeddings (e.g., cosine similarity or Euclidean distance). This loss function encourages the model to capture the causal relationships between events and time-series movements by penalizing counterfactual events that deviate from the causal signal of the ground-truth event.

The combination of diverse counterfactual sampling and causal learning ensures that CAMEF effectively learns the true causal drivers of financial market movements, improving its robustness and predictive power.

% \textbf{Cross-Entropy Loss (Explored but Not Adopted):} We also tested a cross-entropy objective, which aims to maximize the log-likelihood of the ground-truth event being selected over the counterfactual events:

% \begin{equation}
% \mathcal{L}_{\text{Causal-CE}} = - \log \frac{\exp(\mathbf{H}_{\text{final}} \cdot \mathbf{E}^{GT})}{\exp(\mathbf{H}_{\text{final}} \cdot \mathbf{E}^{GT}) + \sum_{i=1}^{|\mathbb{E}^{CF}|} \exp(\mathbf{H}_{\text{final}} \cdot \mathbf{E}^{CF}_i)}.
% \end{equation}

% However, this objective did not perform well in our experiments, where triplet loss served as the primary objective for causal learning.

\noindent \textbf{Total Loss:} The overall training loss for CAMEF is defined as:

\begin{equation}
\mathcal{L}_{\text{Total}} = \mathcal{L}_{\text{Time}} + \mathcal{L}_{\text{Causal-TL}},
\end{equation}

\noindent where \(\mathcal{L}_{\text{Time}}\) is the objective for time series forecasting, as defined in Equation~\ref{eqn:time_objective}.

% \subsection{Training Strategy}

% We adopt a two-phase training strategy:

% \begin{itemize}
%     \item \textbf{Phase 1:} Train only the time-series components (MOMENT, GPT-2, and residual layers) with MSE and MAE objectives, while freezing the RoBERTa encoder. This phase focuses on capturing temporal dependencies accurately.
%     \item \textbf{Phase 2:} Fine-tune all parameters, including RoBERTa, while optimizing both time-series and causal objectives. This phase enables the model to learn cross-modal relations and causal dependencies effectively.

% \end{itemize}

\section{Experiments}

In this section, we evaluate CAMEF by addressing the following key questions: \textbf{RQ1. Accuracy:} How accurately does CAMEF forecast fina ncial market based on events? \textbf{RQ2. Model Effectiveness:} How do different components enhance CAMEF's predictive performance?  \textbf{RQ3. Event Analysis:} Which types of events exhibit stronger influences to financial market?

\subsection{Experimental Settings} 
\label{sec:exp_setting}
\subsubsection{Datasets.}

We utilized the collected event scripts and time-series data as outlined in Sec. \ref{sec:data_acquisition} and detailed in Appendix B. The dataset was divided into training, validation, and testing sets in a 6:2:2 ratio. The training set was used to train the models, while the validation set was used for convergence checking and early stopping to prevent overfitting. The final results, based on the test set, are reported in Table \ref{tab:main_results}.
% For the \textbf{time-series modality}, we evaluate the performance of CAMEF using six multi-modal time-series datasets across three financial categories: \textbf{1) Stock Indices:} S\&P 500 Index (SPX), NASDAQ Index (NDX), Dow Industrial Index (INDU), and Russell 2000 Index (RTY); \textbf{2) US Treasury:} US 1 Month Treasury (USGG1M) and US 5 Year Treasury (USGG5YR); and \textbf{3) Commodity:} Gold Price in US Dollars (XAU). Unlike previous AI-based stock forecasting studies \cite{10.1145/3503161.3548380,ouyang-etal-2024-modal,shah-etal-2023-trillion,Chen_2021_ICCV,pmlr-v162-zhou22g,Jia_Wang_Zheng_Cao_Liu_2024}, we collect and utilize high-frequency time-series data, which provides dense temporal information and closely reflects real-world trading dynamics.

% For the \textbf{textual (event) modality}, we follow prior financial studies and collect event scripts across eight categories, including: \textbf{FOMC Meeting Unemployment Insurance Releases, Initial Unemployment Claims, GDP Advance Releases, CPI Report, and PPI Report}. These scripts were crawled from the official source websites, as detailed in Section \ref{sec:data_acquisition}. We also leverage the LLaMa-3 model to generate 10 counterfactual event scripts for each ground-truth event script, the statistics are included in Table \ref{tab:data_summary}A. We aligned each time-series with the eight categories of events based on their time stamps as explained in Section \ref{sec:definition}.

\subsubsection{Baselines.}
To evaluate our proposed method, we compare it against both \textbf{uni-modal} and \textbf{multi-modal} time series forecasting approaches. For the \textbf{uni-modal baselines}, we considered the traditional yet robust ARIMA model \cite{10.5555/561899}, the linear neural model DLinear \cite{Zeng_Chen_Zhang_Xu_2023}, and several state-of-the-art transformer-based time-series models, including AutoFormer \cite{Chen_2021_ICCV}, FEDformer \cite{pmlr-v162-zhou22g}, and iTransformer \cite{liu2024itransformer}. For the \textbf{multi-modal baselines}, we included TEST \cite{sun2024test} and GPT4MTS \cite{Jia_Wang_Zheng_Cao_Liu_2024}.

\subsubsection{Test Settings}

We evaluate the baselines and CAMEF across three time horizons—short, medium, and long run, to simulate real investment behavior. For each aligned pair of event and time-series data \(([\mathcal{X}_{i-\tau:i+\tau} \mapsto \mathcal{E}_i])\) as defined in Sec.~\ref{sec:definition}, we use the event script (\(\mathcal{X}\)) and the time-series segment preceding the event time point \(i\), i.e., \(\mathcal{X}_{i-\tau:i}\), to forecast the future time-series segment \(\mathcal{X}_{i+1:i+\tau}\). The value of \(\tau\) is adjusted based on the time horizon: we set \(\tau\) to 35, 70, and 140 for short, medium, and long-term forecasts, respectively. These correspond to 175 minutes (about half a trading day), 350 minutes (about one trading day), and 700 minutes (about two trading days).

\begin{table*}[th]
\caption{Ablation Study Results (MSE) Evaluating the CAMEF Model Components at Forecasting Lengths of 35, 70, and 140.}
\label{tab:ablation}
\resizebox{\textwidth}{!}{%
\begin{tabular}{ccccc|rll|lll|lll|lll|lll|lll}
\hline
\multicolumn{1}{l}{} & \multicolumn{1}{l}{} & \multicolumn{1}{l}{} & \multicolumn{1}{l}{} & \multicolumn{1}{l|}{} & \multicolumn{3}{c|}{\textbf{SPX}} & \multicolumn{3}{c|}{\textbf{INDU}} & \multicolumn{3}{c|}{\textbf{NDX}} & \multicolumn{3}{c|}{\textbf{USGG1M}} & \multicolumn{3}{c|}{\textbf{USGG5YR}} & \multicolumn{3}{c}{\textbf{AVERAGE}} \\ \cline{6-23} 
\multicolumn{1}{l}{\multirow{-2}{7.5mm}{\textbf{Textual}}} & \multicolumn{1}{l}{\multirow{-2}{7.5mm}{\textbf{Causal}}} & \multicolumn{1}{l}{\multirow{-2}{7.9mm}{\textbf{\begin{tabular}[c]{@{}l@{}}Feature\\  Fusion\end{tabular}}}} & \multicolumn{1}{l}{\multirow{-2}{9mm}{\textbf{\begin{tabular}[c]{@{}l@{}}GPT2\\ Decoder\end{tabular}}}} & \multicolumn{1}{l|}{\multirow{-2}{12mm}{\textbf{\begin{tabular}[c]{@{}l@{}}Post-\\ Regressor\end{tabular}}}} & \multicolumn{1}{c}{35} & \multicolumn{1}{c}{70} & \multicolumn{1}{c|}{140} & \multicolumn{1}{c}{35} & \multicolumn{1}{c}{70} & \multicolumn{1}{c|}{140} & \multicolumn{1}{c}{35} & \multicolumn{1}{c}{70} & \multicolumn{1}{c|}{140} & \multicolumn{1}{c}{35} & \multicolumn{1}{c}{70} & \multicolumn{1}{c|}{140} & \multicolumn{1}{c}{35} & \multicolumn{1}{c}{70} & \multicolumn{1}{c|}{140} & \multicolumn{1}{c}{35} & \multicolumn{1}{c}{70} & \multicolumn{1}{c}{140} \\ \hline
\ding{55} & \ding{51} & \ding{51} & \ding{51} & \ding{51} & 0.00074 & 0.00340 & 0.00120 & 0.00340 & 0.00330 & 0.01033 & 0.00131 & 0.00097 & 0.00197 & 0.00080 & 0.00179 & 0.00092 & 0.00222 & 0.00276 & 0.00380 & 0.00169 & 0.00199 & 0.00364 \\
\ding{51} & \ding{55} & \ding{51} & \ding{51} & \ding{51} & 0.00068 & 0.00095 & 0.00106 & 0.02939 & 0.00281 & 0.00533 & 0.00064 & 0.00073 & 0.00121 & 0.00065 & 0.00083 & 0.00099 & 0.00160 & 0.00220 & 0.00308 & 0.00659 & 0.00170 & 0.00233 \\
\ding{51} & \ding{51} & \ding{55} & \ding{51} & \ding{51} & 0.00080 & 0.00079 & 0.00110 & 0.01173 & 0.00391 & 0.00581 & 0.00069 & 0.00073 & 0.00168 & 0.00047 & 0.00046 & 0.00216 & 0.00251 & 0.00306 & 0.00426 & 0.00324 & 0.00212 & 0.00300 \\
\ding{51} & \ding{51} & \ding{51} & \ding{55} & \ding{51} & 0.00073 & 0.00067 & 0.00114 & 0.26768 & 0.72387 & 0.18091 & 0.00062 & 0.00069 & 0.00158 & 0.00043 & 0.02110 & 0.75057 & 0.00220 & 0.00309 & 0.00566 & 0.05433 & 0.14593 & 0.03801 \\
\ding{51} & \ding{51} & \ding{51} & \ding{51} & \ding{55} & 0.00662 & 0.03911 & 0.00979 & 0.50841 & 0.57007 & 0.22267 & 0.00774 & 0.00852 & 0.00950 & 0.28932 & 0.02110 & 0.75057 & 0.56705 & 0.06188 & 0.08251 & 0.27583 & 0.14014 & 0.21501 \\ \hline
\multicolumn{5}{c|}{\textbf{Full CAMEF Model}} & \multicolumn{1}{l}{\textbf{0.00048}} & \textbf{0.00064} & \textbf{0.00107} & \textbf{0.00253} & \textbf{0.00250} & \textbf{0.00393} & \textbf{0.00054} & \textbf{0.00058} & \textbf{0.00101} & \textbf{0.00028} & \textbf{0.00044} & \textbf{0.00049} & {\color[HTML]{000000} \textbf{0.00132}} & \textbf{0.00207} & \textbf{0.00224} & \textbf{0.00104} & \textbf{0.00124} & \textbf{0.00174} \\ \hline
\end{tabular}
}
\end{table*}

\begin{table}[th]
\caption{Ablation Study Results on Different Type of Events on S\&P500 Index}
\label{tab:ablation2}
\resizebox{\columnwidth}{!}{%
\begin{tabular}{c|ll|ll|ll}
\hline
\multirow{2}{*}{\textbf{Event Type}} & \multicolumn{2}{c|}{\textbf{\begin{tabular}[c]{@{}c@{}}Forecasting Length = 35\end{tabular}}} & \multicolumn{2}{c|}{\textbf{\begin{tabular}[c]{@{}c@{}}Forecasting Length = 70\end{tabular}}} & \multicolumn{2}{c}{\textbf{\begin{tabular}[c]{@{}c@{}}Forecasting Length = 140\end{tabular}}} \\
 & \multicolumn{1}{c}{\textbf{MSE}} & \multicolumn{1}{c|}{\textbf{MAE}} & \multicolumn{1}{c}{\textbf{MSE}} & \multicolumn{1}{c|}{\textbf{MAE}} & \multicolumn{1}{c}{\textbf{MSE}} & \multicolumn{1}{c}{\textbf{MAE}} \\ \hline
\textbf{\begin{tabular}[c]{@{}c@{}}Unemployment\\ Insurance\end{tabular}} & 0.0004870 $\downarrow$ & 0.0159497 $\uparrow$ & 0.0005199 $\downarrow$ & 0.0157778 $\downarrow$ & 0.0006948 $\downarrow$ & 0.0190141 $\downarrow$ \\
\textbf{\begin{tabular}[c]{@{}c@{}}Employment\\ Situation\end{tabular}} & 0.0003923 $\downarrow$ & 0.0142684 $\downarrow$ & 0.0004612 $\downarrow$ & \textbf{0.0154431} $\downarrow$ & 0.0013767 $\uparrow$ & 0.0218677 $\uparrow$ \\
\textbf{GDP Adcance} & 0.0006203 $\uparrow$ & 0.0175397 $\uparrow$ & 0.0005897 $\downarrow$ & 0.0175464 $\downarrow$ & 0.0011554 $\uparrow$ & 0.0217548 $\uparrow$ \\
\textbf{FOMC Minutes} & \textbf{0.0003401} $\downarrow$ & \textbf{0.0127694} $\downarrow$ & \textbf{0.0004448} $\downarrow$ & 0.0170094 $\downarrow$ & \textbf{0.0006433} $\downarrow$ & \textbf{0.0185657} $\downarrow$ \\
\textbf{CPI Report} & 0.0005645 $\uparrow$ & 0.0160723 $\uparrow$ & 0.0010660 $\uparrow$ & 0.0212536 $\uparrow$ & 0.0008187 $\downarrow$ & 0.0197824 $\downarrow$ \\
\textbf{PPI Report} & 0.0005275 $\uparrow$ & 0.0148434 $\downarrow$ & 0.0008054 $\uparrow$ & 0.0201844 $\uparrow$ & 0.0017646 $\uparrow$ & 0.0251856 $\uparrow$ \\ \hline
\multicolumn{1}{l|}{\textbf{Full Selection}} & 0.0004886 & 0.0152405 & 0.0006478 & 0.0178691 & 0.0010756 & 0.0210284 \\ \hline
\end{tabular}
}
\end{table}

\subsubsection{Implementation Overview}

For the single-modality approaches (ARIMA, DLinear, AutoFormer, FEDformer, iTransformer, and PatchTST), we tested two methods: (1) training the models on continuous historical time-series data and testing on aligned event-based time-series segments from the test set, and (2) training the models directly on event-based time-series segments, and also test on the aligned event-based time-series segments from the test set. The second approach produced more accurate results, and these are the results presented in this paper. Specifically, for each aligned event data pair \(([\mathcal{X}_{i-\tau:i+\tau} \mapsto \mathcal{E}_i])\), single-modality approaches use only the time-series segment preceding the event, \(\mathcal{X}_{i-\tau:i}\), to train the models and forecast the subsequent segment, \(\mathcal{X}_{i+1:i+\tau}\); and testing follows the same approach. Detailed settings for each model are provided in Appendix C.

For the multi-modality approaches (GPT4MTS, TEST, and CAMEF), both the event script \(\mathcal{E}_i\) and the time-series segment preceding the event, \(\mathcal{X}_{i-\tau:i}\), are used as input to train the models to forecast the post-event time-series segment, \(\mathcal{X}_{i+1:i+\tau}\). Implementation details, including model configurations and training settings, are explained in Appendix C.

\subsection{Experimental Results for Event-Driven Time-Series Forecasting (RQ1)}

Table \ref{tab:main_results} presents the forecasting results for the five datasets across short, medium, and long forecasting horizons. CAMEF outperformed other models in 24 out of 30 settings, achieving first-place rankings, and ranked second in the remaining 6 settings. Specifically, CAMEF demonstrated the best performance across all forecasting lengths for the Stock Market Indices (SPX, INDU, and NDX), except for short-horizon forecasting on INDU, highlighting its effectiveness in event-driven stock market forecasting. For treasury bonds, CAMEF achieved the best results across all forecasting lengths for the 1-month treasury bond (USGG1M) and ranked second for the 5-year treasury bond (USGG5YR). The slightly lower performance on USGG5YR suggests that long-run treasury bonds may be less sensitive to event-driven factors and more influenced by historical trends.

Compared to single-modality models (e.g., DLinear, Autoformer, FEDformer, PatchTST, and iTransformer), CAMEF achieved an average MSE reduction of 62.55\% relative to the best-performing single-modality model, iTransformer. Among multi-modality models, CAMEF surpassed TEST, the second-best performer, with an average MSE reduction of 33.55\%. 

These results highlight three key insights: (1) effectively leveraging multi-modality information provides significant performance gains, particularly for SPX, INDU, NDX, and USGG1M; (2) transformer-based methods consistently outperform classical models such as ARIMA; and (3) CAMEF’s superior training and feature fusion strategies establish it as the most effective method for event-driven financial forecasting tasks.

% \begin{itemize}
%     \item \textbf{Conventional Baseline (ARIMA):} \\
%     ARIMA, a statistical model focused on linear dependencies, performs relatively well on simpler assets, like Gold Price (XAUUSD), but shows higher MSE on volatile indices (e.g., S\&P500, Nasdaq100). Its single-modality nature limits its ability to capture complex patterns in financial time series.

%     \item \textbf{Transformer-Based Models (Autoformer, FEDformer, iTransformer, PatchTST):} \\
%     Transformer models improve over ARIMA by capturing long-range dependencies, with iTransformer and PatchTST achieving lower MSE on assets such as Nasdaq100 and S\&P500. However, lacking multi-modal information, these models show limitations, particularly for longer forecasting horizons.

%     \item \textbf{TEST (Text-Enhanced Transformer):} \\
%     TEST incorporates text data but shows inconsistent performance, particularly for short-term forecasts. This suggests that while text features contribute, their integration with time-series data needs refinement to enhance forecast stability.

%     \item \textbf{Proposed Multi-Modality Model (CAMEF):} \\
%     Our CAMEF model, combining RoBERTa-based text embeddings with MOMENT/GPT2-based time-series encoding, achieves the lowest MSE across nearly all assets. CAMEF’s robust multi-modal fusion and causal learning yield strong improvements on high-volatility indices and Treasury Bonds, highlighting its adaptability and performance advantage over single-modality baselines.
% \end{itemize}

\subsection{Ablation Studies on Model Components (RQ2)}

To evaluate the effectiveness of each component in CAMEF, we conduct comprehensive ablation studies on Textual Modality, Causal Learning, Feature Fusion, GPT2 Decoder, and the Post-Regressor. Based on the full CAMEF model, we remove the corresponding neural layers, such as the RoBERTa encoder for textual modality or the causal learning component, to assess their individual contributions. The ablation results are presented in Table \ref{tab:ablation}, where the left part of the table uses \ding{51} or \ding{55} to indicate whether the specific component is included or excluded in the test. From the results, three key findings are: (1) The full CAMEF model achieves the best performance across all datasets, demonstrating the critical importance of utilizing both textual and time-series modalities; (2) Causal learning provides incremental improvements, confirming its value in capturing cause-effect relationships within the data; (3) The proposed feature fusion layers and GPT2 decoder effectively integrate and leverage multi-modality features, significantly enhancing the model's ability to decode time-series data. These findings underscore the necessity of each component in achieving optimal performance for event-driven financial forecasting tasks.

% \begin{figure*}[ht]
%     \centering
%     % First subplot
%     \begin{subfigure}[b]{0.32\textwidth} % Adjust width as needed
%         \centering
%         \includegraphics[width=\textwidth]{figures/[CPI]trend_following_strategy_with_thresholds_analysis.png} % Replace with actual path
%         \caption{Potential Cumulative Return on CPI Report Released in Mar. 2024}
%         \label{fig:CPI_return}
%     \end{subfigure}
%     \hfill
%     % Second subplot
%     \begin{subfigure}[b]{0.32\textwidth}
%         \centering
%         \includegraphics[width=\textwidth]{figures/[FOMC]trend_following_strategy_with_thresholds_analysis.png} % Replace with actual path
%         \caption{Potential Cumulative Return on FOMC Minute Released in Oct. 2023}
%         \label{fig:FOMC_return}
%     \end{subfigure}
%     \hfill
%     % Third subplot
%     \begin{subfigure}[b]{0.32\textwidth}
%         \centering
%         \includegraphics[width=\textwidth]{figures/[UI]trend_following_strategy_with_thresholds_analysis.png} % Replace with actual path
%         \caption{Potential Cumulative Return on Unemployment Insurance Released in Feb. 2024}
%         \label{fig:UI_return}
%     \end{subfigure}

%     \caption{Potential Cumulative Returns on S\&P500 Index for CPI (Mar. 2024), FOMC (Oct. 2023), and Unemployment Insurance (Feb. 2024) Events Using Trend-Following Strategy Based on Predictions from the CAMEF Model.}
%     \label{fig:Ptential_Returns}
% \end{figure*}

\subsection{Ablation Studies on Different Type of Events (RQ3)}

Table \ref{tab:ablation2} shows the predictive performance of different events on the S\&P500 Index. \textbf{FOMC Minutes} achieve the lowest MSE and MAE, confirming their critical importance for market prediction. \textbf{Unemployment Insurance Claims} and \textbf{Unemployment Situation Reports} also come with lower errors than the full selection, however the latter becomes less effective at long forecasting length. In contrast, \textbf{CPI} and \textbf{PPI Reports} show weaker predictive power, with PPI yielding the highest errors and CPI improving slightly at long forecasting length. These results emphasize the importance of FOMC and unemployment-related events for financial forecasting.

\subsection{Parameter Sensitivity Analysis}

\begin{figure}[t]
\centering
\begin{subfigure}{0.48\linewidth}
    \centering
    \includegraphics[width=\linewidth]{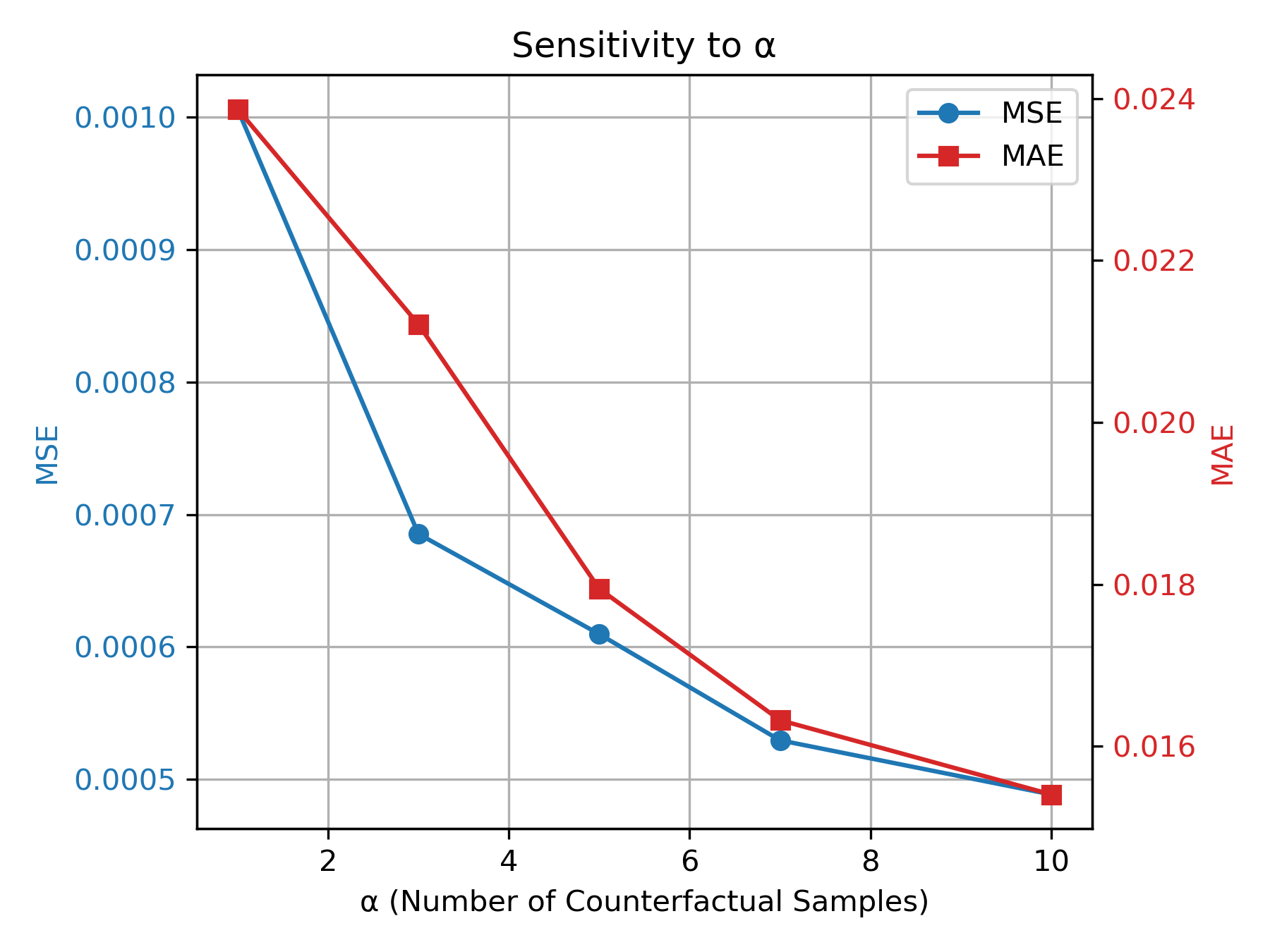}
    \caption{Effect of counterfactual sample size ($\alpha$)}
    \label{fig:sensitivity-alpha}
\end{subfigure}
\hfill
\begin{subfigure}{0.48\linewidth}
    \centering
    \includegraphics[width=\linewidth]{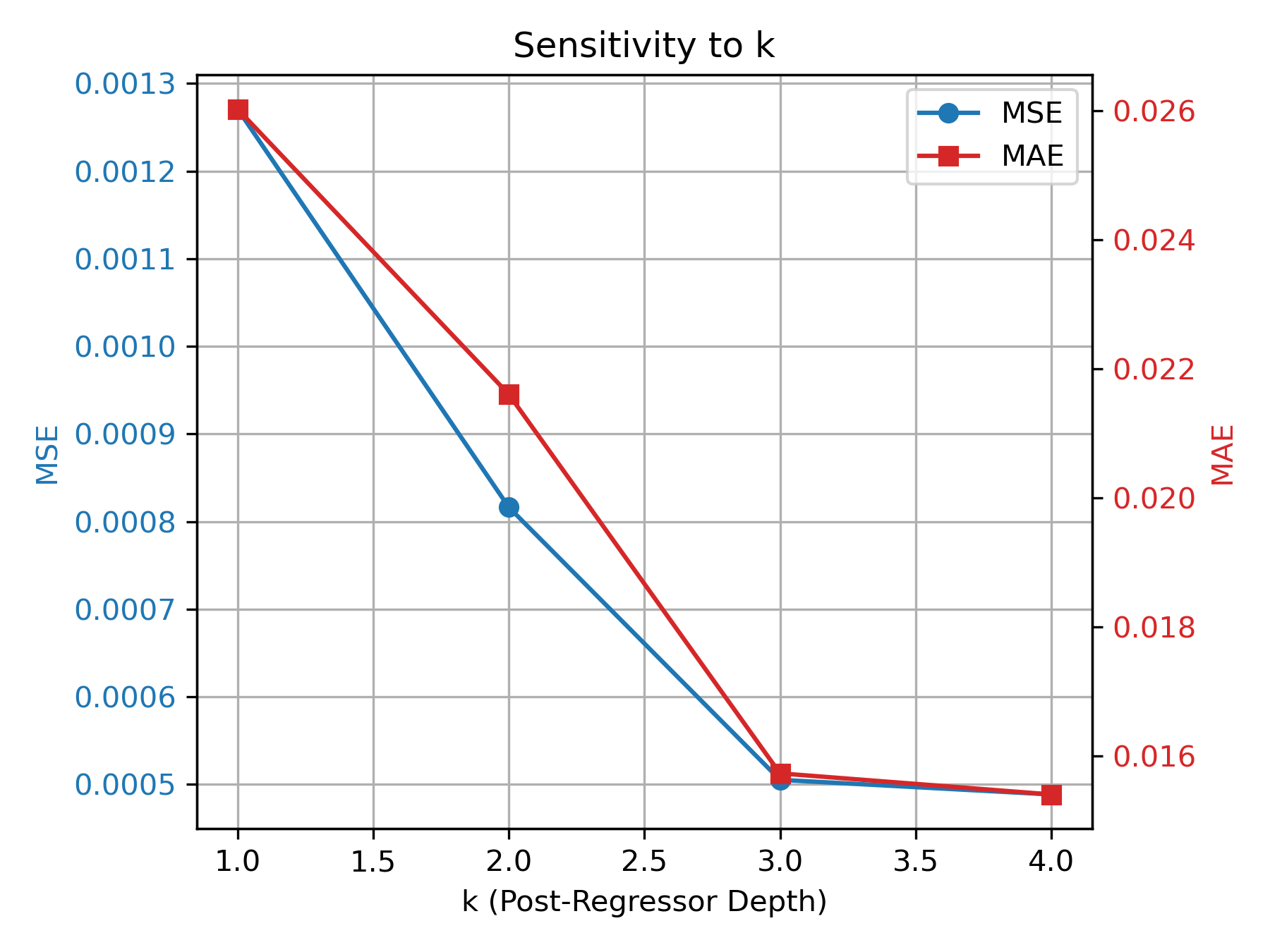}
    \caption{Effect of post-regressor depth ($k$)}
    \label{fig:sensitivity-k}
\end{subfigure}
\caption{Sensitivity analysis of CAMEF on the SP500 dataset with respect to key hyperparameters.}
\label{fig:sensitivity}
\end{figure}

We examine the impact of two key hyperparameters in CAMEF on SPX dataset with 35 predictive length: the number of counterfactual samples ($\alpha$) and the depth of the post-regressor network ($k$). The number of GPT-2 layers ($l$) remains fixed, as we adopt the pre-trained model without modification. As shown in Figure~\ref{fig:sensitivity}, increasing $\alpha$ enhances forecasting accuracy through stronger contrastive supervision, with performance gains saturating around $\alpha = 10$. Similarly, increasing $k$ improves decoding capacity, though with diminishing returns beyond $k = 4$. These trends suggest that CAMEF performs robustly across a range of configurations, benefiting from moderate complexity increases without overfitting.

\section{Conclusion and Future Work}

This paper proposed \textbf{CAMEF}, a multi-modality model for event-driven financial forecasting, which integrates effective causal learning and an LLM-based counterfactual event augmentation strategy. Alongside the model, we introduced a novel synthetic dataset comprising 6 types of salient macroeconomic event scripts, their counterfactual samples, and high-frequency time-series data for 5 key financial assets, aligned with real-world investment practices. Extensive experiments demonstrated CAMEF's superior predictive performance compared to prior deep time-series and multi-modality methods. Ablation studies testified the importance of causal learning and other designed components of CAMEF. Additionally, it is found that FOMC and unemployment-related events provided the most predictive value among the tested event types.

For future work, we plan to leverage advanced LLMs for enhanced textual encoding to extract deeper semantic information, refine the cross-modality causal inference mechanisms, and expand the dataset to include additional event types, such as political events and corporate market-sensitive news.

\section*{Acknowledgements}

This work was supported in part by the National Natural Science Foundation of China (NSFC) under Grant Nos. 62402396 and 72471197. We would like to thank the anonymous reviewers for their valuable comments and suggestions that helped improve the quality of this paper.

\newpage

\bibliographystyle{ACM-Reference-Format}
\bibliography{references}

%%
%% If your work has an appendix, this is the place to put it.
\appendix

\section{Prompt Templates for Counterfactual Events Generations}
\label{sec_app:prompt_templates}
\subsection{Summarization Prompt Template}
Due to the script’s length, we firstly prompt LLaMA-3 summarizes each chunk of an event script, followed by a second prompt to generate the full summary. Below is the chunk-level prompt, which instructs the model to summarize within a word limit. Text in ``{ }'' indicates input variables.

\begin{tcolorbox}[colback=gray!5!white, colframe=gray!75!black, title=Prompt Template for Event Script Chunk Summarization]
You are given chunk \{\textit{chunk\_idx}\} of a \{\textit{text\_type}\} report. Your task is to generate a summary within \{\textit{number\_of\_words}\} words.\\
The content of chunk \{\textit{chunk\_idx}\} is as follows:\\
\{\textit{original\_text}\}\\
Please provide a concise summary, while keep the key variables:
\end{tcolorbox}

\noindent The second prompt combines the chunk summaries as input to generate a final summary for the entire event script:

\begin{tcolorbox}[colback=gray!5!white, colframe=gray!75!black, title=Prompt Template for Final Summarization]
You are given \{\textit{chunk\_num}\} summaries of different chunks from a \{\textit{text\_type}\} report. Your task is to generate an overall summary within \{\textit{number\_of\_words}\} words.\\
The chunk summaries are as follows:
\{\textit{chunk\_summaries}\}\\
Please provide a comprehensive summary of the entire report, while keep the key variables:
\end{tcolorbox}

% \vspace{-1em}

\subsection{Sentiment Analytical Prompt Template}

For sentiment analysis, the prompt asks the model to rate an event scirpt's sentiment from 0 (negative) to 10 (positive) and explain the rating. Below is the prompt template, with ``{ }'' denoting input variables.

\begin{tcolorbox}[colback=gray!5!white, colframe=gray!75!black, title=Sentiment Analysis Prompt]
Please analyze the sentiment of the following \{\textit{text\_type}\} summary and rate it on a scale from 0 to 10, where:\\
0 = Extremely Negative; 1 = Strongly Negative; 2 = Very; Negative; 3 = Moderate Negative; 4 = Slightly Negative; 5 = Neutral; 6 = Slightly Positive; 7 = Moderate Positive; 8 = Very Positive; 9 = Strongly Positive; 10 = Extremely Positive
\{\textit{text\_type}\} summary: \{\textit{text}\}\\
Output the sentiment analysis as:\\
Sentiment rating: (0 to 10), Explanation:
\end{tcolorbox}

\subsection{Counterfactual Event Generation Prompt Template}

To generate counterfactual versions of a text with different sentiment levels, we use a prompt that instructs the model to modify key facts and information to align with a target sentiment rating. The model is provided with the original sentiment rating and sentiment description and is asked to adjust the text to reflect a specified target sentiment rating. Below is the prompt, where the text within ``\{ \}'' indicates the input variables:

\begin{tcolorbox}[colback=gray!5!white, colframe=gray!75!black, title=Counterfactual Text Generation Prompt]
The original text has been identified with a sentiment rating of \{\textit{current\_sentiment\_rating}\} (\{\textit{current\_sentiment}\}).\\
Your task is to generate a counterfactual version of the text that aligns with a sentiment rating of \{\textit{target\_sentiment\_rating}\} (\{\textit{target\_sentiment}\}) by modifying the key facts and information to reflect the specified target sentiment score about the economy, while keep the overall format and the sentiment-neural content unchanged.\\
Original text: \{\textit{original\_text}\}\\
Counterfactual text with a sentiment rating of \{\textit{target\_sentiment\_rating}\} (\{\textit{target\_sentiment}\}): 
\end{tcolorbox}

\section{Implementation Details of Dataset Collection and Preprocessing}
\label{sec_app:implementaion_dataset}

\subsection{Dataset Collection}

To construct the CAMEF dataset, raw macroeconomic event data are crawled from official government sources (Tab. \ref{tab:data_summary}) in various formats (PDF, HTML, TXT). We utilize the following Python libraries:

\begin{itemize}
\item \textbf{Requests} – to send HTTP requests for archive access;
\item \textbf{Selenium} – to locate and download files via HTML headers;
\item \textbf{BeautifulSoup} – to parse HTML and extract file links.
\end{itemize}

The dataset covers six event types: FOMC minutes, CPI, PPI, unemployment insurance, unemployment rate, and GDP advance releases.

\subsection{Preprocessing}

Raw files are converted into a unified text format. \texttt{PyPDF2} and \texttt{pdfplumber} extract content from PDFs, while \texttt{BeautifulSoup} parses HTML using depth-first traversal. TXT files are used as-is. We convert all the tables into structured text using the delimiter “|”, as shown:

\begin{verbatim}
<Table>
Header 1 | Header 2 | Header 3
----
Cell 1  | Cell 2    | Cell 3
</Table>
\end{verbatim}

This preprocessing ensures clean, consistent, machine-readable data, supporting robust analysis and model training.

\section{Implementation Details}

Single-modality baselines (ARIMA, DLinear, AutoFormer, FEDFormer, iTransformer, PatchTST) were adapted from Time-Series-Library to the event-driven setting, using pre-event segments to predict 35-, 70-, and 140-step trends. All models were trained for 10 epochs with batch size 32. 

For multi-modality baselines, we used TEST and a re-implemented GPT4MTS (with LongFormer replacing BERT). Both were aligned to the same forecasting horizons and training settings.

CAMEF followed a two-stage pipeline: MOMENT was pre-trained on time series, then the full model fine-tuned on aligned text–series pairs. MOMENT and RoBERTa (excluding final layer) were frozen, while GPT2 and fusion layers were trainable, enabling robust multimodal forecasting.

\subsection{Implementation of Baseline Models}
Baseline models were developed with objectives distinct from our event-driven forecasting approach, as single-modality methods are primarily designed for continuous time-series forecasting. To enable a fair comparison, we adapted these models to align with an event-driven context, where past time-series data preceding an event is used to forecast trends within a defined forecasting horizon.

\textbf{Single-Modality Models:} For time-series-based single-modality models, including \textbf{DLinear, AutoFormer, FEDFormer, iTransformer, and PatchTST}, we utilized the open-source library \textbf{Time-Series-Library}\footnote{\url{https://github.com/thuml/Time-Series-Library}}. Time-series segments were extracted with lengths covering both the input and forecasting periods surrounding each event, ensuring alignment with the respective event's time step. The input segment represents the historical trend information, while the forecasting period is treated as "unseen" data to be predicted. Input and forecasting lengths were consistently set to 35, 70, and 140 time steps across all experiments, as described in Sec.~\ref{sec:exp_setting}. All models were trained for 10 epochs, with a batch size of 32. These settings ensured convergence of the loss function and maintained consistency with CAMEF's training configuration.

\textbf{Multi-Modality Models:} For multi-modality methods, including \textbf{TEST} and \textbf{GPT4MTS}, we followed implementation strategies specific to each model.

\begin{itemize}
    \item \textbf{TEST:} We followed the instructions provided in the open-source TEST repository\footnote{\url{https://github.com/SCXsunchenxi/TEST}}, which involved two training stages. In the first stage, the encoder was trained by selecting 10 prototype words based on GPT vocabulary clustering (as instructed in the repository) to align textual representations with time-series data. The second stage involved training the time-series decoder, where input and forecasting steps were set to 35, 70, and 140, similar to the single-modality models. The batch size was set to 7, and training epochs were kept at 10.

    \item \textbf{GPT4MTS:} As the official implementation of GPT4MTS was unavailable, we re-implemented the model based on its original paper. Instead of using BERT \cite{devlin-etal-2019-bert} as the textual encoder, we employed LongFormer \cite{beltagy2020longformerlongdocumenttransformer}, which is better suited for encoding longer contexts, as our texts tend to be relatively lengthy. Input time-series segments were used to forecast the time-series data at the forecasting lengths of 35, 70, and 140. We kept training epochs at 10 and the batch size at 7 to ensure consistency with other baselines.
\end{itemize}

\subsection{Implementation of CAMEF}

\begin{table}[t]
\centering
\caption{Learning Rates for CAMEF Components}
\label{tab:learning_rates_CAMEF}
\resizebox{0.5\columnwidth}{!}{  
\begin{tabular}{l|c}
\hline \hline
\textbf{Model Component}               & \textbf{Learning Rate} \\ \hline
MOMENT Model                           & \(1 \times 10^{-6}\)   \\
RoBERTa Model                          & \(5 \times 10^{-7}\)   \\
GPT2 Model                             & \(1 \times 10^{-5}\)   \\
Embedding Layer                        & \(1 \times 10^{-5}\)   \\
Residual Layer                         & \(1 \times 10^{-5}\)   \\ 
Fusion Layer                           & \(5 \times 10^{-7}\)   \\
Output Layer                           & \(1 \times 10^{-5}\)   \\ \hline \hline
\end{tabular}
}
\end{table}

\noindent The batch size is set to 10, and the training epochs are set to 10, consistent with the baseline models.

The implementation of \textbf{CAMEF} involves two distinct training stages to ensure effective learning of both time-series and textual features, while leveraging pre-trained components:

\begin{enumerate}
    \item \textbf{Pre-training MOMENT:} In the first stage, we pre-train the MOMENT model using the time-series segments from the training data. This step focuses on learning  representations of the past time-series patterns.
    \item \textbf{Training Entire CAMEF:} In the second stage, we train the entire CAMEF model using both event scripts and their corresponding aligned time-series segments. During this stage, certain parameters are frozen to retain the pre-trained knowledge, while others remain open for fine-tuning:
    \begin{itemize}
        \item \textbf{Frozen Parameters:} All parameters of the MOMENT model and RoBERTa (except for the last hidden layer) are frozen. Additionally, the token embedding layer of GPT2 in the decoder is frozen to preserve its pre-trained representations.
        \item \textbf{Trainable Parameters:} The remaining components of GPT2, the last hidden layer of RoBERTa, and other components such as the embedding, residual, fusion, and output layers are open for training.
    \end{itemize}
\end{enumerate}

The hidden sizes of the different components of the CAMEF model are detailed in Sec.~\ref{sec:model_architecture}. To optimize the model effectively, we adopt component-specific learning rates as listed in Table~\ref{tab:learning_rates_CAMEF}.

% BLOCK THE RIGHT COLUMN WITH PHANTOM BOX
\hfill
\phantom{\parbox[t]{0.48\textwidth}{This phantom box blocks the right column.}}

\end{document}